\documentclass[journal]{IEEEtran}
\ifCLASSINFOpdf
\else
\fi

\usepackage{cite}
\usepackage{graphicx} 
\usepackage{epstopdf}
\usepackage{amsfonts}
\usepackage{multicol}  
\usepackage{multirow}  
\usepackage{color}
\usepackage{subfigure}
\usepackage{amsmath}
\usepackage{soul}%
\definecolor{ColorName}{rgb}{0,0,0}
\usepackage{xcolor}
\usepackage[normalem]{ulem} % use normalem to protect \emph
\usepackage[framemethod=tikz]{mdframed}
\usepackage{threeparttable}%

\usepackage{url}

\usepackage{hyperref}
\begin{document}
\title{DASGIL: Domain Adaptation for Semantic and Geometric-aware Image-based Localization}

\author{Hanjiang~Hu, Zhijian~Qiao, Ming~Cheng, Zhe~Liu, Hesheng~Wang$^{*}$,~\IEEEmembership{Senior~Member,~IEEE} %Weidong~Chen,~\IEEEmembership{Member,~IEEE}%
\thanks{This work was supported in part by the Natural Science Foundation of China under Grant U1613218, U1913204, 62073222 and 61722309. Corresponding Author: Hesheng Wang. }% <-this % stops a space
\thanks{H. Hu, Z. Qiao, M. Cheng, and H. Wang are with Department of Automation, Insititue of Medical Robotics, Key Laboratory of System Control and Information Processing of Ministry of Education, Key Laboratory of Marine Intelligent Equipment and System of Ministry of Education, Shanghai Jiao Tong University, Shanghai 200240, China. H. Wang is also with Beijing Advanced Innovation Center for Intelligent Robots and Systems, Beijing Institute of Technology, China. Z. Liu is with the Department of Computer Science and Technology, University of Cambridge, United Kingdom. } }

%\author{Michael~Shell,~\IEEEmembership{Member,~IEEE,}
        %John~Doe,~\IEEEmembership{Fellow,~OSA,}
        %and~Jane~Doe,~\IEEEmembership{Life~Fellow,~IEEE}% <-this % stops a space
%\thanks{M. Shell was with the Department
%of Electrical and Computer Engineering, Georgia Institute of Technology, Atlanta,


\markboth{IEEE Transactions on Image Processing,~Vol.XXX, No.XXX, July~2020}%
{HU \MakeLowercase{\textit{et al.}}: DASGIL: Domain Adaptation for Semantic and Geometric-aware Image-based Localization}



\maketitle

\begin{abstract}
Long-Term visual localization under changing environments is a challenging problem in autonomous driving and mobile robotics due to season, illumination variance, \textit{etc}. Image retrieval for localization is an efficient and effective solution to the problem. In this paper, we propose a novel multi-task architecture to fuse the geometric and semantic information into the multi-scale latent embedding representation for visual place recognition. To use the high-quality ground truths without any human effort, \textcolor{ColorName}{the effective multi-scale feature discriminator is proposed for adversarial training to achieve the domain adaptation from synthetic virtual KITTI dataset to real-world KITTI dataset.} The proposed approach is validated on the Extended CMU-Seasons dataset \textcolor{ColorName}{and Oxford RobotCar dataset} through a series of crucial comparison experiments, where our performance outperforms state-of-the-art baselines for retrieval-based localization \textcolor{ColorName}{and large-scale place recognition} under the challenging environment. 
\end{abstract}

\begin{IEEEkeywords}
Visual localization, image retrieval, representation learning, domain adaptation.
\end{IEEEkeywords}

% \fi
\IEEEpeerreviewmaketitle

\section{Introduction}
\label{intro}
\IEEEPARstart{V}{isual} localization plays an essential role in mobile robots and outdoor self-driving vehicles \cite{gao2018ldso,bescos2018dynaslam}, especially for long-term \textcolor{ColorName}{Simultaneous Localization and Mapping (SLAM)} systems, in which environmental factors including illumination, weather and seasonal changes have significant influence on the precision of visual localization\cite{sattler2018benchmarking}. \textcolor{ColorName}{Image-based localization is an efficient way to retrieve target image from database given the query image across different environments. Once the place recognition presents the coarse results, the high-precision 6-DoF camera pose could be regressed with the retrieved initial value\cite{sarlin2019coarse}.}

	Traditional local descriptors like SIFT, ORB, \textit{etc.} work satisfactorily in image matching without much variance of environmental conditions. However, due to the dependence on image pixels, these man-made local features are not robust under drastically varying conditions. Global feature shows impressive advantages on visual localization, \textit{e.g.} \textcolor{ColorName}{Vector of Locally Aggregated Descriptors (VLAD)}-like feature \cite{jegou2010aggregating,arandjelovic2016netvlad}  and DenseVLAD \cite{torii201524} gives impressive performance on long-term image-based localization. In the meanwhile, with the great development of deep neural networks, especially \textcolor{ColorName}{Convolutional Neural Networks (CNNs)} in computer vision, the deep and dense features extracted from CNN have been used in image-based localization in changing environments \cite{sunderhauf2015performance}, showing deeper features are more robust to change of view point. As feature goes deeper, the semantic information has been more extracted and therefore could be used to generate depth map\cite{eigen2014depth,wang2015designing} or semantic segmentation map \cite{long2015fully,zhao2017pyramid}. 
% 	Besides, recent work \cite{hu2019retrieval} trains shallow feature to be domain-invariant. 
	
%Recent work \cite{hu2019retrieval} presents a pipeline for training in an self-supervised manner to learn domain-invariant feature from images across multiple domains.
	
	\textcolor{ColorName}{Since the  deeper feature is more semantic, the perceptual information could improve the robustness to the changing pixels caused by environmental variance through higher-level tasks, like monocular depth prediction and semantic segmentation.} Recent work \cite{piasco2019learning,piasco2020improving} leverages the geometric information with auxiliary depth map for image-based localization. Besides,  \cite{larsson2019fine,larsson2019cross} introduce the semantic segmentation for the improvement of visual localization. \textcolor{ColorName}{However, it is effort-costly and time-consuming to obtain the depth and semantic segmentation maps for real-world images. Consequently, synthetic data and domain adaptation have drawn significant attention in recent years, aiming to use  the high-quality groundtruth with the least human effort. But there is a huge domain gap between the virtual and real dataset, which is challenging for the domain adaptation\cite{nath2018adadepth,zhao2019geometry}.} 
	
	To efficiently leverage the geometric and semantic information for visual localization with zero labor costs, we propose DASGIL, a novel domain adaptation with semantic and geometric information for image-based localization. The proposed method adopts the one-encoder-two-decoder multi-task architecture, fusing geometric and semantic information to multi-scale latent features through the shared Fusion Feature Extractor. The fused features of both virtual and real images follow the same distribution in multi-layer-feature adversarial training \textcolor{ColorName}{through the novel Flatten and Cascade Discriminator}, adapting from synthetic images to real-world images. Based on the fused multi-scale features, metric learning for place recognition is accomplished through multi-scale triplet loss for metric learning. For the experiments, we train the model on Virtual KITTI 2 dataset but test it on Extended CMU-Seasons dataset and \textcolor{ColorName}{Oxford RobotCar dataset} for retrieval-based localization and place recognition, and our results are better than state-of-the-art methods under various regional environments, vegetation conditions and weather conditions. 
	
	 In summary, our work makes the following contributions:
	\begin{itemize}
		\item We propose a novel and state-of-the-art approach, DASGIL, fusing semantic and geometric information into latent features through a multi-task architecture of depth prediction and semantic segmentation. 
		\item \textcolor{ColorName}{A novel domain adaptation framework is introduced using multi-scale feature discriminator through adversarial training from synthetic to real-world dataset for representation learning.} 
		\item Multi-scale metric learning for place recognition is adopted through multi-layer triplet loss and features from different scales are applied in retrieval process as well.
		\item A series of comparison experiments have been conducted to validate the effectiveness of every proposed module in DASGIL. And our approach outperforms state-of-the-art image-based localization baselines on the Extended CMU-Seasons dataset \textcolor{ColorName}{and Oxford RobotCar dataset} though only supervisely trained on Virtual KITTI 2 dataset.
	\end{itemize}
	
%	We structure the rest of this paper as follows. Section. \ref{related} analyzes the related work. The architecture of DASGIL and the  pipeline for image-based localization are introduced in
%	Section. \ref{archi}  and Section. \ref{pipe}, respectively.  Section. \ref{expe} introduces the training details, the experimental results and ablation study.	Finally, in Section. \ref{conc} we draw our conclusions for this work.

\section{Related Work}

\label{related}
	\subsection{Domain Adaptation for Segmentation and  Depth Prediction}
	In the applications of visual perception in autonomous driving and mobile robotics, monocular depth prediction and semantic segmentation are significantly important and have been constantly studied in recent years. Man-made descriptors have been used in the traditional robotic applications \cite{gu2020active,han2020vision}. Since deep convolutional neural networks (DCNN) boost the performance of visual perception, enormous studies show promising results for monocular depth estimation \cite{eigen2014depth,wang2015designing, godard2019digging} and semantic segmentation \cite{long2015fully}, where fully convolutional neural networks (FCNNs) are introduced  to enable end-to-end training.
	
	However, the groundtruth of depth map and segmentation map in outdoor scenarios are often time-consuming and expensive to obtain, which constrains the development and performance of supervised learning methods. Some unsupervised or weakly-supervised methods are proposed for depth prediction \cite{godard2017unsupervised,watson2019self} and semantic segmentation \cite{larsson2019cross,larsson2019fine}, with the assistance of left-right consistency \cite{godard2017unsupervised} or ego-motion pose constraint\cite{zhou2017unsupervised} for depth prediction and  image level tags or 2D-2D points\cite{larsson2019cross} for semantic segmentation. 
	
	Besides, the virtual synthetic datasets \cite{gaidon2016virtual,ros2016synthia, cabon2020virtual} are developed to deal with such issue as well, where the image sequences are under changing environments with perfectly high-quality depth map and segmentation map as groundtruth. However, as the models are trained on these virtual datasets, there occurs a concern about the generalization ability to the real-world images with the domain gap.
	
	Domain adaptation refers to the generalization ability to a new different dataset for a model trained on one dataset. Some previous works focus on domain-invariant deep latent feature learning \cite{long2013transfer} in which \cite{long2013transfer} mitigates the discrepancy between source and target domain to make the feature distribution identical by minimizing Maximum Mean Discrepancy (MMD). While adversarial loss \cite{ghifary2016deep} is also commonly used to decrease the discrepancy, through training the discriminator to discriminate the representation feature from source or target domain.
	
	As for the domain adaptation in the task of depth prediction \cite{atapour2018real,nath2018adadepth,zhao2019geometry} and semantic segmentation \cite{sankaranarayanan2018learning, murez2018image}, lots of work focus on image translation \cite{atapour2018real,zhao2019geometry}, which largely depends on the performance of translation \cite{zhu2017unpaired} and is hard to cover all the changing environments, so they not suitable for place recognition in various conditions. \cite{zheng2018t2net} proposes to train the mid-feature to be consistent between two domains while it only copes with single-layer feature, lacking the adaptation of multi-scale features.

	\subsection{Long-term Place Recognition and Localization}
	Outdoor visual place recognition has been studied for many years and could be directly used for visual localization in autonomous driving or loop closure detection of SLAM, in which the most similar images are retrieved from database for query images. Traditional local feature descriptors are aggregated for image retrieval \cite{galvez2012bags,cummins2008fab}, and have successfully addressed most cases of loop closure detection in real-time \textcolor{ColorName}{Visual SLAM (VSLAM)}\cite{mur2017orb,milford2012seqslam} without huge environmental changes. VLAD \cite{jegou2010aggregating} is the most successful man-made feature for place recognition and it has been extended to different versions. NetVLAD\cite{arandjelovic2016netvlad} extract deep features through VLAD-like network architecture.  DenseVLAD \cite{torii201524} presents retrieval results through extracting multi-scale SIFT descriptor with VLAD\cite{jegou2010aggregating} under drastic perceptual changes. \textcolor{ColorName}{Also, Regional Maximum Activations of Convolutions (R-MAC) \cite{tolias2015particular} is also used for image retrieval which acts as an effective pooling for each feature map patch.}
		
	Since convolutional neural networks (CNNs) has successfully addressed many tasks in computer vision, long-term visual place recognition and localization has developed significantly aided with CNN. Image translation-based methods seem to be the most direct way to solve the problem, where images are transfered across different domains based on generative adversarial networks (GANs) \cite{goodfellow2014generative}. \textcolor{ColorName}{Night-to-day translation \cite{anoosheh2019night,zheng_2020_ECCV} similarly translates night-images to day-images to overcome the adverse condition for robust perception and localization.} Jenicek \textit{et al.} \cite{jenicek2019no} proposes to use U-Net to obtain photometric normalization image and then find deep embedding for retrieval. However, generalization ability is limited for translation-based methods because the accuracy of image-level retrieval largely depends on the quality of translated image compared to latent-feature retrieval of ours.
	
	\textcolor{ColorName}{To cope with the challenging perceptual change, unlike the direct point-wise feature learning in point cloud based place recognition \cite{liu2019lpd}, many recent works follow the pipeline of learning the robust deep representation through neural networks together with semantic \cite{xu2018unsupervised,benbihi2020image}, geometric\cite{piasco2019learning,piasco2020improving}, context-aware information\cite{chen2018learning,xin2019localizing}, \textit{etc}. However, these works need auxiliary information which is effort-cost to obtain in most cases. Instead of explicit image translation, latent feature learning is promising for image retrieval \cite{hu2019retrieval,zhou2020cross,tang2020adversarial}.} Consequently, domain adaptation at the level of feature map seems prospective for place recognition assisted with geometric and semantic information.

	\section{DASGIL Architecture}
	\label{archi}
	\begin{figure*}[thpb]
	\centering
	%\framebox
	%{
	\includegraphics[scale=0.55]{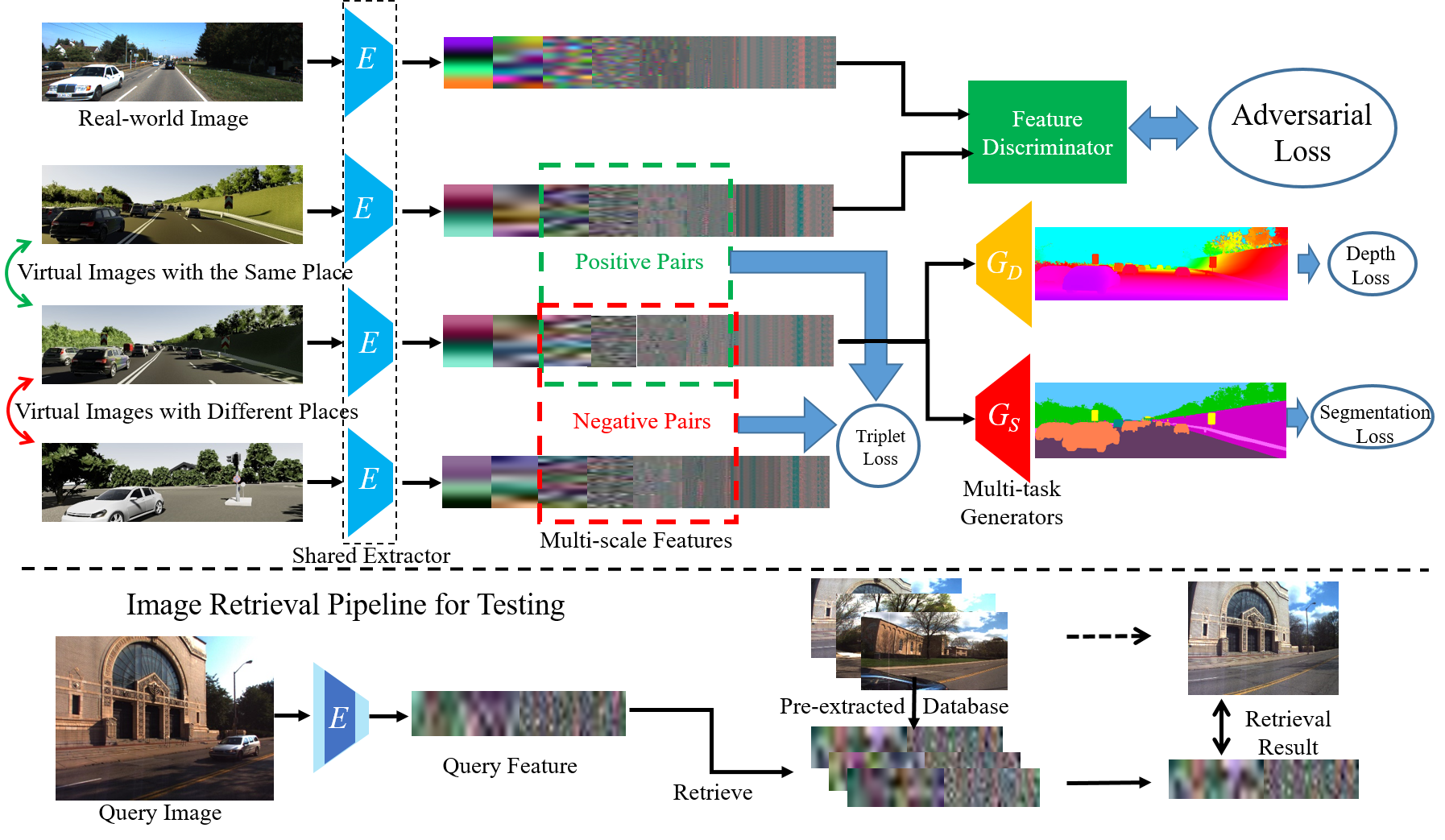}
%	\vspace{-0.5cm}
	%}
	
	\caption
	{
		The overview of the proposed DASGIL approach and the image retrieval pipeline are shown above. Note that only the virtual depth map and segmentation map groundtruth are used for supervised training with little human cost. The multi-layer feature is visualized through PCA dimension reduction on the feature map from every layer.
	}
%	\vspace{-0.5cm}
	\label{overview}
\end{figure*}
	\subsection{Architecture Overview}

	As shown in Figure. \ref{overview}, our proposed DASGIL adopts one-encoder-two-decoders architecture, including one shared fusion feature extractor $ E $ and two map generators $ G_{S}, G_{D} $ for semantic segmentation and depth prediction, respectively. The extracted multi-scale features from  $ E $ are used to generate target depth map and segmentation map   given the virtual image input $ I_{V} \sim p_{V}(I) $. To diminish the domain gap between synthetic images and real-world images, the adversarial training is incorporated through the multi-scale feature discriminator $D$, resulting in the same distribution of multi-scale features from the inputs of both the  real images $ I_{R} \sim p_{R}(I) $ and virtual images $ I_{V} \sim p_{V}(I)$. \textcolor{ColorName}{Note that the visualization of multi-scale feature map is based on PCA which reduces the channel to RGB dimension for each pixel on the feature map, showing the distribution on the feature map. The denser the visualization is, the large the feature map is.}

	\subsection{Fusion Feature Extractor}
	\label{extractor}
	    To extract geometric and semantic information from the input RGB image ($I$), we use a shared encoder ($E$) to accomplish this task. Since the depth map and segmentation map are both based on the extracted features through $E$, the extractor fuses geometric information and semantic information into the multi-scale features. Besides, deep features of multiple scales are extracted from $E$ through all the convolution layers for the skip connection to decoders as U-Net-like models \cite{ronneberger2015u,isola2017image} do. \textcolor{ColorName}{Specifically, the feature extractor involves eight CNN layers with the skip connection to the depth and segmentation map generator for each layer.} This structure instructs the model to obtain and use different levels of features containing geometric and semantic information, which assists the the generation of depth map and segmentation map.
	    
	\begin{figure}[thpb]
		\centering
		%\framebox
		%{
		\includegraphics[scale=0.45]{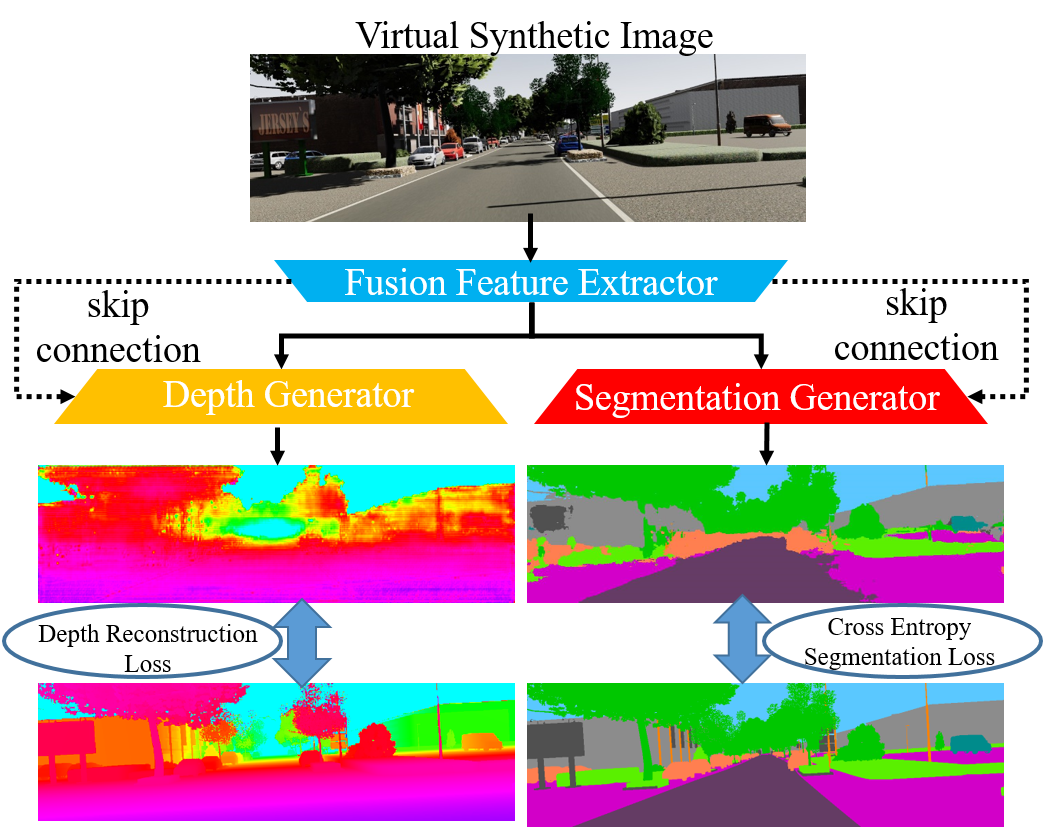}
		%	\vspace{-0.5cm}
		%}
		
		\caption
		{
			The extractor and multi-task generators for depth map and segmentation map are shown above, where the depth reconstruction loss and cross entropy segmentation loss are used for virtual synthetic images input.
		}
		%	\vspace{-0.5cm}
		\label{multitask}
	\end{figure}

	\subsection{Depth Map And Segmentation Map Generator}
	\label{discriminator}
	Two map generation networks ($G_D$, $G_S$) with the same structure are used to generate depth map and segmentation map, respectively.  In order to instruct the model to generate depth map and segmentation map based on the multi-scale features extracted from $E$, we use skip-connections from $E$ to $G_D$ and $G_S$, as Figure. \ref{multitask} shows. Notice that the segmentation map and depth map are decoded from the same fusion features extracted from the shared encoder $E$. The  depth map and segmentation map generated are shown in Figure. \ref{multitask}.

    \subsection{Multi-scale Discriminator}

\begin{figure}[thpb]
   	\centering
   	
   	\begin{mdframed}[hidealllines=true]%,backgroundcolor=blue!20]%,
   		%\lipsum[2]
   		\includegraphics[scale=0.32]{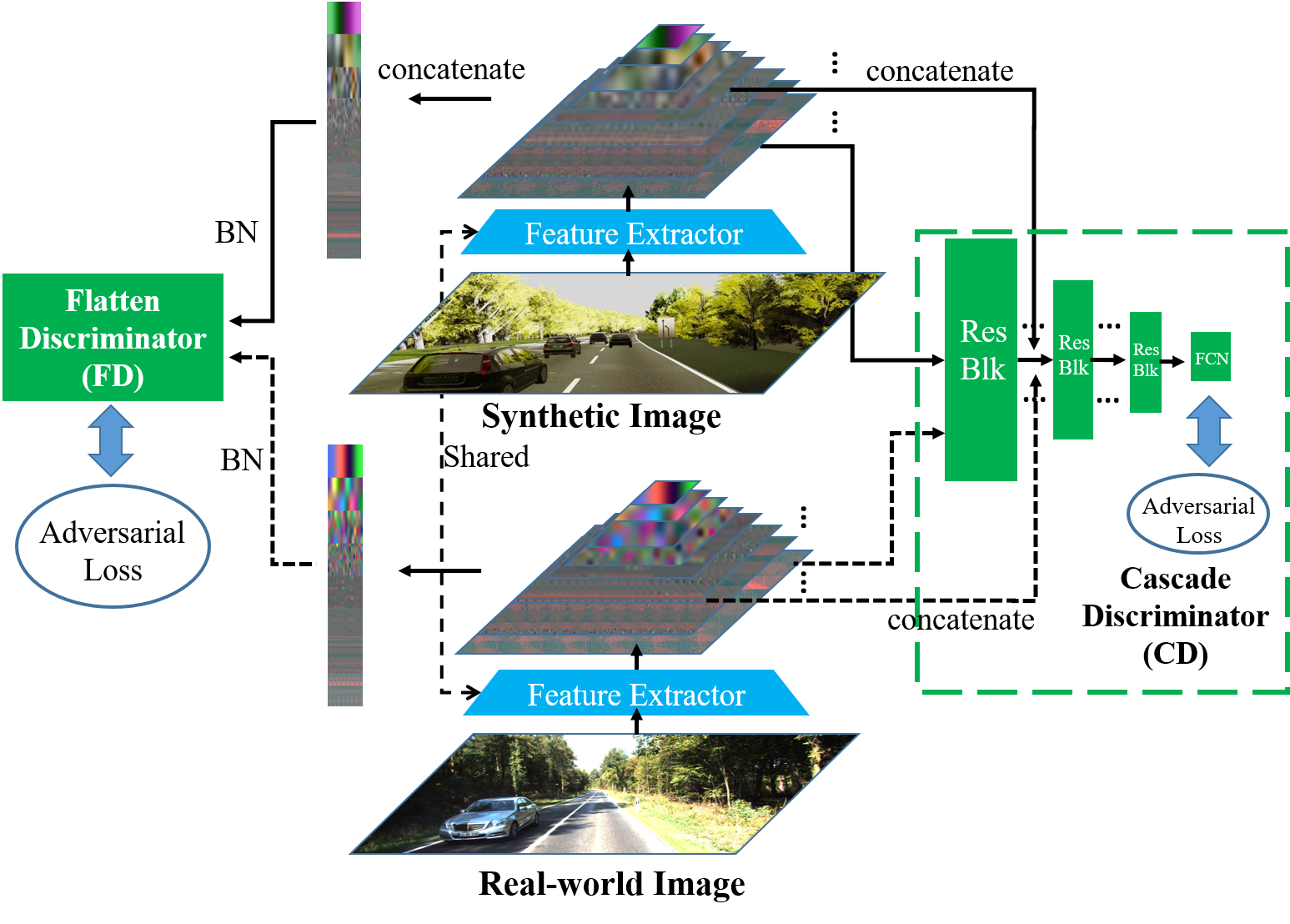}
   	\end{mdframed}
   	%\framebox
   	%{
   	
   	%	\vspace{-0.5cm}
   	%}
   	
   	\caption
   	{
   		\textcolor{ColorName}{The Flatten Discriminator (FD) is shown on the left while the Cascade Discriminator (CD) is shown on the right. For FD, the dense multi-layer features are extracted from $E$ and then concatenated to feed the multi-scale feature discriminator for adversarial training given synthetic and real-world images. For CD, the next deeper feature is concatenated to the output of ResBlock with the input of the previous feature and the final classification is obtained through a fully connected layer. The multi-layer feature is visualized through PCA dimension reduction on the feature map from every layer.}
   	}
   	%	\vspace{-0.5cm}
   	\label{multigan}
   \end{figure}

    Since the feature extractor $ E $ is trained on the virtual synthetic images $I_{V}$ while tested on real-world images $I_{R}$ for image localization, the extracted mid-features fused with geometric and semantic information must be distribution-consistent for both virtual images and real-world images. For the domain adaptation from synthetic domain to real-world domain, the adversarial training strategy is adopted in the multi-scale latent embedding space. Different from existing domain adaptation work \cite{atapour2018real, zhao2019geometry} which build the discriminator at image or single-scale feature level, we build the discriminator at the level of features with multiple scales, as shown in Figure. \ref{multigan}.

    \textcolor{ColorName}{A simple multi-scale discriminator is implemented as follows. The multi-scale features are concatenated and then go through a batchnorm layer before fed into the feature discriminator $D$. The proposed feature discriminator consists of three fully connected layers as a two-class classifier to determine whether the latent feature is from real-world domain $R$ or virtual domain $V$. The structure is denoted as Flatten Discriminator (FD), as shown in Figure. \ref{multigan}.}
    
    \textcolor{ColorName}{However, since features  are extracted to different extents at different layers, it my be not efficient to discriminate all the concatenated features directly. In order to discriminate shallow and deep features differently across domains, we further propose a Cascade Discriminator (CD). For each feature map at a specific layer, the shallow feature map is input to a CNN and the output is connected to the next deeper feature map, resulting in the final classification of real or virtual image.}
    
\begin{figure}[htbp]
	
	\centering
	
	\subfigure[]{
		\begin{minipage}[t]{0.31\linewidth}
			\centering
			\includegraphics[width=\linewidth]{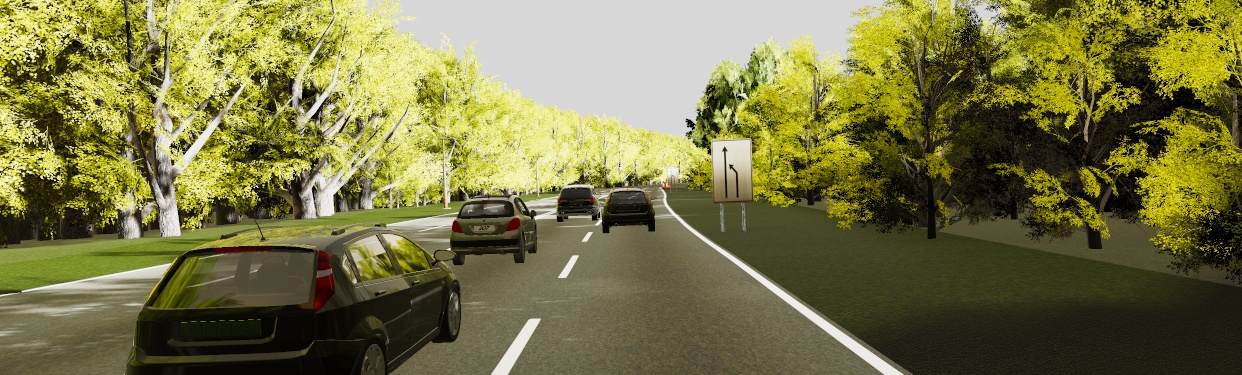}\\
%			\vspace{0.1cm}
			\includegraphics[width=\linewidth]{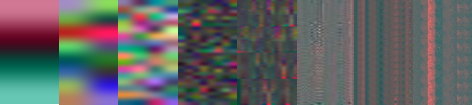}\\
				\includegraphics[width=\linewidth]{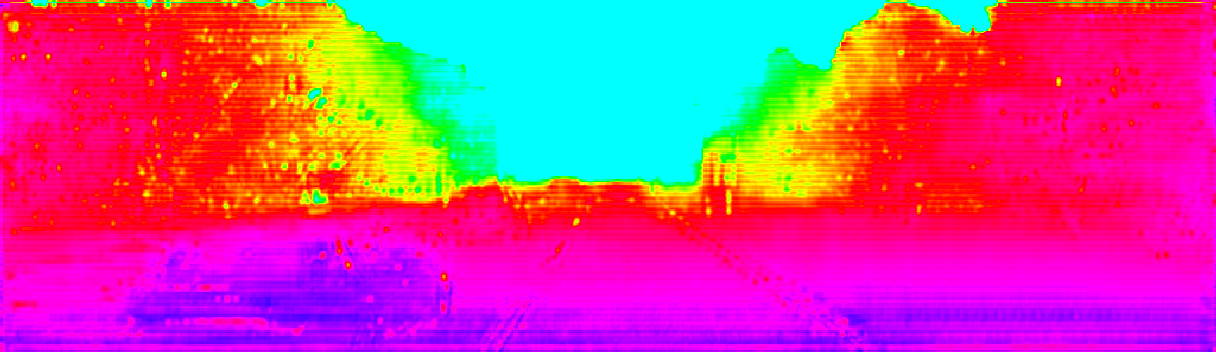}\\
					\includegraphics[width=\linewidth]{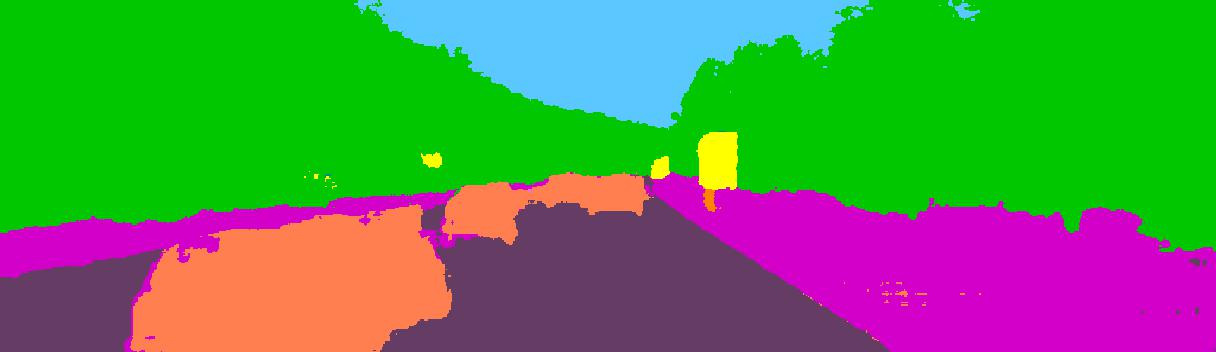}\\
%			\vspace{0.1cm}
			%\caption{fig1}
		\end{minipage}%
	}%
	\subfigure[]{
			\begin{minipage}[t]{0.31\linewidth}
				\centering
				\includegraphics[width=\linewidth]{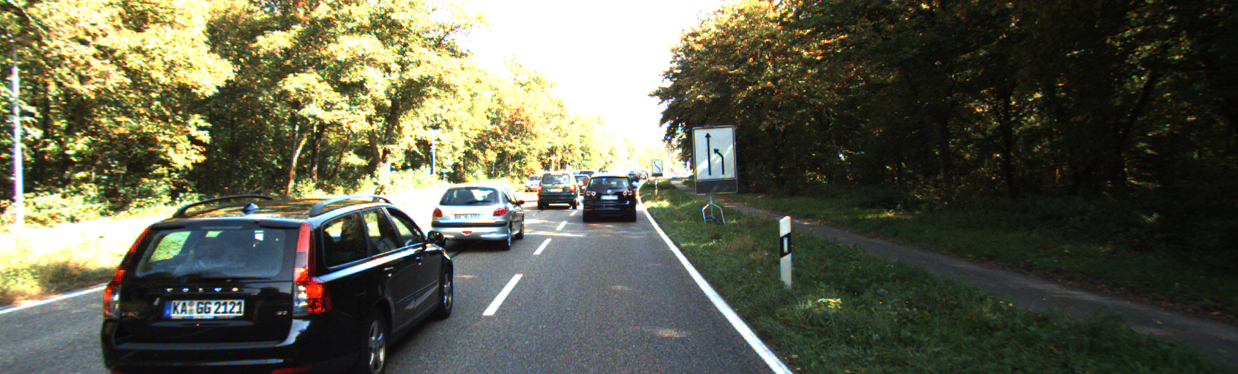}\\
				%			\vspace{0.1cm}
				\includegraphics[width=\linewidth]{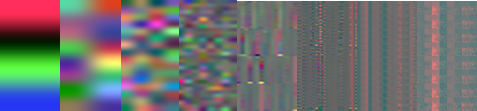}\\	\includegraphics[width=\linewidth]{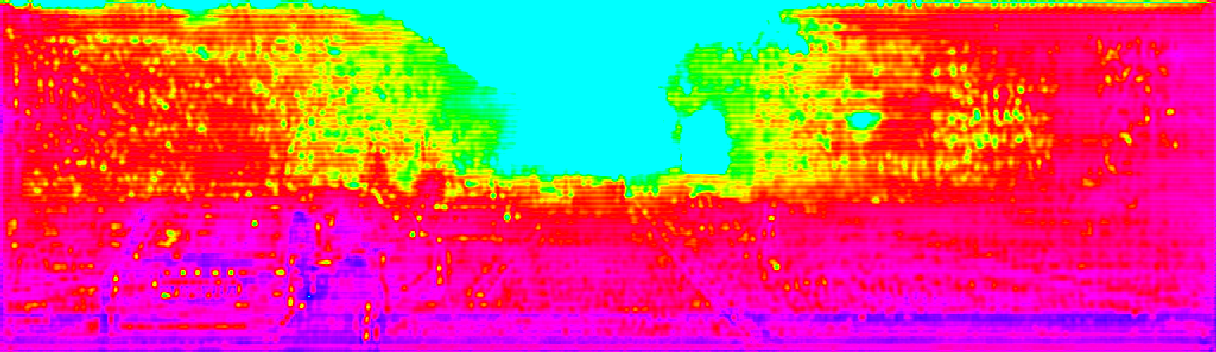}\\
						\includegraphics[width=\linewidth]{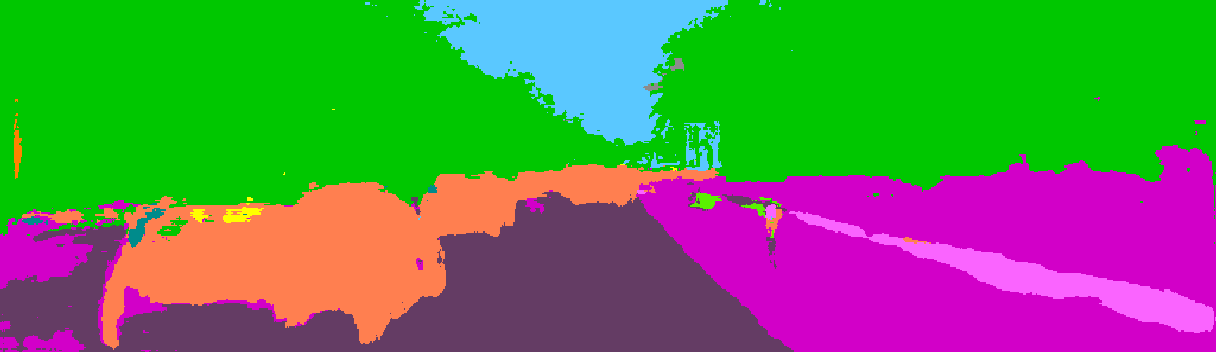}\\
				%			\vspace{0.1cm}
				%\caption{fig1}
			\end{minipage}%
			
		}%
		\subfigure[]{
		\begin{minipage}[t]{0.31\linewidth}
			\centering
			\includegraphics[width=\linewidth]{2011_09_29_drive_0004_sync_image_02_data_0000000142.png}\\
			%			\vspace{0.1cm}
			\includegraphics[width=\linewidth]{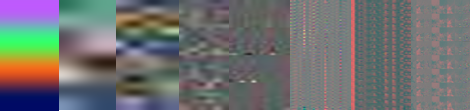}\\	\includegraphics[width=\linewidth]{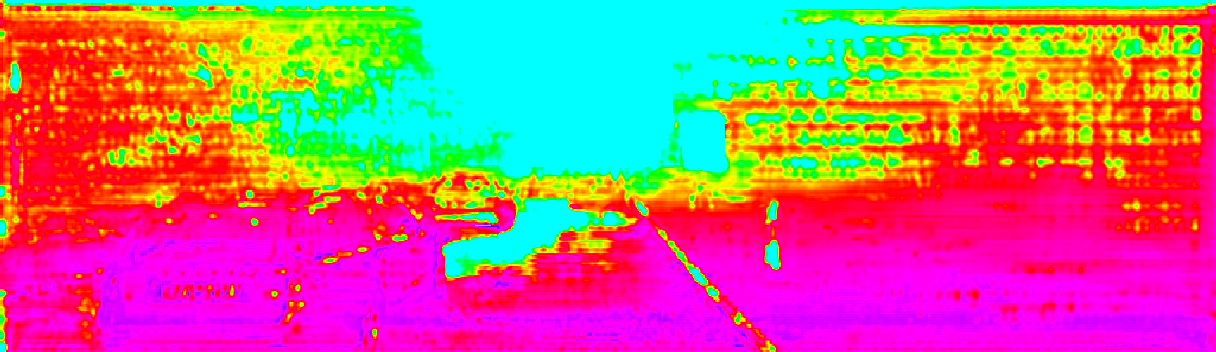}\\	\includegraphics[width=\linewidth]{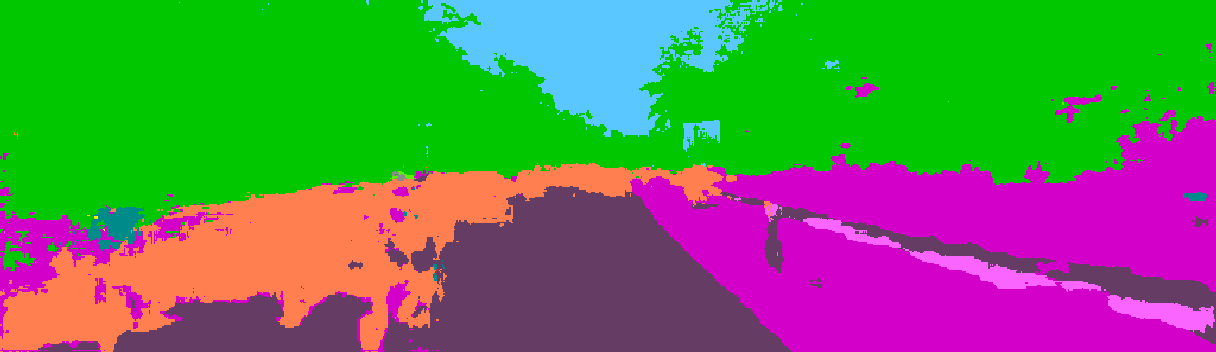}\\
			%			\vspace{0.1cm}
			%\caption{fig1}
		\end{minipage}%
		
	}%

	\centering
	%	\vspace{-0.2cm}
	\caption{The effectiveness and necessity of adversarial training are illustrated in the figure. From top to bottom, each row represents the original RGB image, multi-scale feature visualization from PCA dimension reduction, predicted depth map and segmentation map respectively. Column (a) show the results for virtual KITTI images, while Column (b) and (c) are for real-world KITTI images. Results of Column (b) are obtained from the model trained with adversarial loss from Flatten Discriminator while results of (c) are without adversarial training. It could be seen that the quality of depth segmentation maps from (b) are significantly improved through adversarial training compared with (c), and the visualized features of (a) and (b) are almost under the same distribution due to the effective domain adaptation.}
	%	\vspace{-0.5cm}
	\label{ganmatters}
\end{figure}

    \textcolor{ColorName}{These two multi-layer discriminator structures allow the model to recognize $R$ and $V$ from multiple levels, enabling the model to better utilize the fusion information extracted. The effectiveness of FD and CD multi-layer discriminator is validated in \ref{ablation_section} and also shown in Figure. \ref{ganmatters}.} 
   
    \section{DASGIL Pipeline for Image-based Localization}
    \label{pipe}
    For the image retrieval of localization, the multi-task architecture model is trained to learn the fusion feature representation, incorporating the geometric and semantic information into the latent representation. The learning pipeline and training losses are introduced first, and then the image-based localization process is presented.
    
    \subsection{Domain Adaptation for Multi-task Training}
    
    The overall architecture is designed for multi-task, \textit{i.e.} generating both semantic segmentation map and depth map at the same time. The reconstructed depth loss and segmentation loss are introduced to instruct the extractor $ E$ and generators$ G_{S},G_{D}$ to learn latent embedding features and generate the two target maps for virtual input images in Figure. \ref{multitask}. Besides, the adversarial loss is to assure that both of the real-world images and virtual images could extract the same-distributed features through the same extractor $ E$ in Figure. \ref{multigan}.
    
    \subsubsection{\textbf{Multi-scale Depth Reconstruction Loss}}
	Inspired by \cite{godard2019digging}, we construct the multi-scale reconstruction loss for the generation of depth map given the virtual images $I_V\sim p_{V}(I)$. 
	After the forward propagation of $E, G_D$, we utilize the results obtained from multiple layers of $G_D$ and then resize the ground truth of depth map $Depth_{GT}$ into corresponding sizes. Then we compute $L1$ Loss between the features of $G_D$ and ground truth $Depth_{GT}$ at each level. Finally we add the losses together to get the reconstruction loss for depth map.
	\begin{eqnarray}
	\label{multi-scale_depth}
	 \mathcal{L}_{D} = \mathbb{E}_{I_{V}\sim p_{V}(I)}[ \sum_{i}^{n}||G_{D}(E(I_{V}))_{i} - Depth_{GT_{i}}||_1 ]
	 \end{eqnarray}
	where $i$ refers to the $i^{th}$ layer of extractor $E$ and generator $G_D$, and $n$ is the total layers involved, while $Depth_{GT}$ with subscript represents the ground truths which have been resized to the size of $i^{th}$ layer. \textcolor{ColorName}{Note that the involved layers are with large scale and resolution, which are from $ 4^{th} $ to $ 1^{st} $ layer.}

% 	 \mathcal{L}_{S_R} = \mathbb{E}_{I_{V} \sim p_{V}(I)} [\sum_{i = 1}^{n}||D_{D}(E(I_{V}))_{i} - L_{Si}||_1]

    \subsubsection{\textbf{Cross Entropy Segmentation Loss}}
    As for segmentation map, we use cross entropy loss to train the segmentation generator model. The generation of segmentation map is the classification at pixel level for the discrete classification instead of continuous regression. \textcolor{ColorName}{Unlike the image forward propagation for depth reconstruction, we apply score regression and the soft-max layer for the outputs only at the last layers of generator $G_S$, which is for the classification not the regression problem. The cross entropy loss is adopted for every pixel.}
    \textcolor{ColorName}{\begin{eqnarray}
    	\label{multi-scale_seg}
	 \mathcal{L}_{S} = \mathbb{E}_{I_{V} \sim p_{V}(I)}[-\sum_{c=1}^{M} Seg_{GT_c}\log(p_{c})]
	 \end{eqnarray}}
	 \textcolor{ColorName}{where $p_c$ denotes the probability for class $c$ pixel-wisely while $Seg_{GT_c}$ is the ground truth one-shot label for class $c$, and there are $M$ classes in all.}
    
    \subsubsection{\textbf{Multi-scale Feature Adversarial Loss}}
	
	Instead of constructing GAN loss at image level which has been done by most of the existing work in domain adaptation, we construct GAN loss at multiple scales of features and force them to have consistent distribution so as to make full use of the features extracted through each layer in $E$ as shown in Figure. \ref{multigan}. Because the features of \textit{all} layers in $E$ are skip-connected to $G_D$ and $G_S$, it is of great necessity to consider all features when constructing our discriminator and thus the distribution of generated depth map and segmentation map is consistent between virtual and real-world domains. 

	Suppose the image from virtual domain as $ I_{V} \sim p_{V}(I) $, image from real-world domain as $ I_{R} \sim p_{R}(I) $, \textcolor{ColorName}{we use the LSGAN \cite{mao2017least}} form instead of the original form \cite{goodfellow2014generative}.
	\textcolor{ColorName}{
	\begin{eqnarray}
	\label{GAN-D}
	&\mathcal{L}_{Dis} = \mathbb{E}_{ I_{V} \sim p_{V}(I),I_{R} \sim p_{R}(I)}[0.5 \times (D(E(I_{V}))^{2}\\ \notag
		&+D(E(I_{R})-1)^{2})]  \\
	\label{GAN-G}
	&\mathcal{L}_{Gen} = \mathbb{E}_{ I_{V} \sim p_{V}(I)}[0.5 \times D(E(I_{V})-1)^{2}]
	\end{eqnarray}}
	where $D$ is the multi-layer feature discriminator described in \ref{discriminator} and $E$ is the fusion feature extractor described in \ref{extractor}. \textcolor{ColorName}{During the adversarial training, $\mathcal{L}_{Dis}$ and $\mathcal{L}_{Gen}$ are optimized separately.}	We validate the effectiveness of our multi-layer adversarial loss in Section. \ref{ablation_section}.
    
    \subsection{Representation Learning for Place Recognition}
    While the multi-task model is trained to fuse the geometric and semantic information into the latent embedding feature, the key point of retrieval-based localization is to learn the robust representations for database and query images. Also for the deep metric learning task, the fusion triplet loss is calculated using concatenated features extractor from $ E$ in multiple scales.
	
	\begin{figure}[thpb]
		\centering
		%\framebox
		%{
		\includegraphics[scale=0.3]{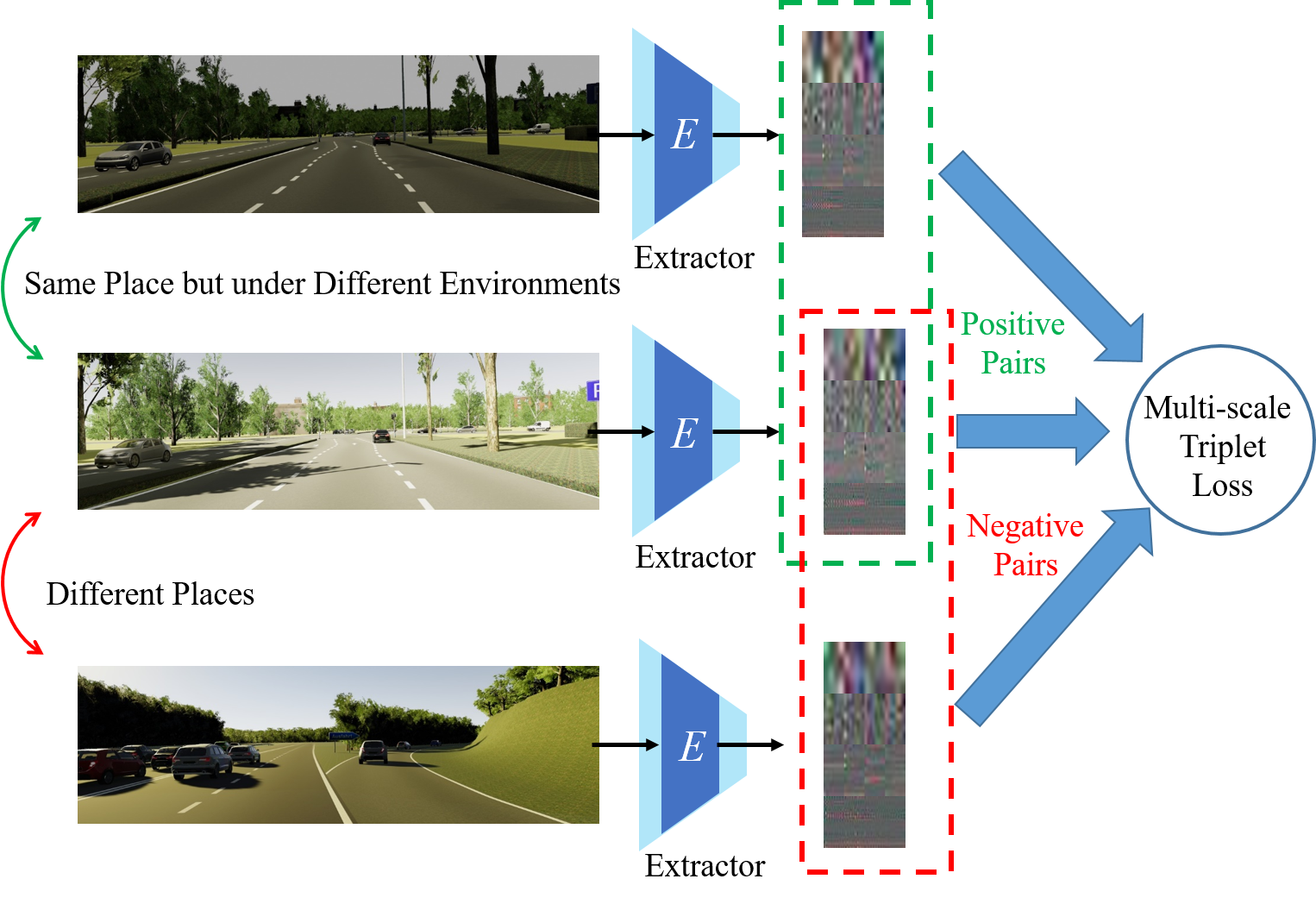}
		%	\vspace{-0.5cm}
		%}
		
		\caption
		{
			The multi-scale triplet loss is calculated through middle features from multiple layers of $E$ given positive image pairs under the same place and negative image pairs under different places.
		}
		%	\vspace{-0.5cm}
		\label{tripletfig}
	\end{figure}
	
	\subsubsection{\textbf{Multi-scale Fusion Triplet Loss}}
	\label{triplet_loss}
	
    We propose a \textit{Multi-scale Fusion Triplet Loss} to instruct the model to learn specific representation, as shown in Figure. \ref{tripletfig}. Different from \cite{piasco2019learning} which generates depth maps and extracts geometric information explicitly through another depth encoder before calculating the triplet loss, we improve this strategy by fusing both geometric information and semantic information explicitly for metric learning.
	A triplet loss involves an anchor virtual image ($q_i \sim q_V(I)$), a positive sample ($q_{i+}$) representing the same scene as the anchor and a negative sample ($q_{i-}$) which is unrelated to the anchor image. The triplet loss ($L_{t}(q_i, q_{i+}, q_{i-}, m)$) is shown in the formula below:
	\begin{eqnarray}
	\label{triploss}
		\mathcal{L}_{T} = \mathbb{E}_{ q_i, q_{i+}, q_{i-} \sim p_{V}(I)}[max(0, 1 - \\ \notag
		\frac{||q_i - q_{i-}||_2}{margin + ||q_i - q_{i+}||_2})]
	\end{eqnarray}
	where $m$ represents the margin value how the distance of negative pairs is larger than that of positive pairs and would be a constant.
	
	Considering the impact of features at different levels on the final task, retrieval-based localization, we improve the formula (\ref{triploss})
	and propose a multi-scale triplet loss in formula (\ref{multi_triploss})
	\textcolor{ColorName}{\begin{eqnarray}
%	\begin{split}
	\label{multi_triploss}
	&\mathcal{L}_{mul-T} = \mathbb{E}_{ q_i, q_{i+}, q_{i-} \sim p_{V}(I)}[\sum_{l = L_m}^{L_k} \mathcal{L}_T^l] \\ \notag
	&=\mathbb{E}_{q \sim p_{V}(I)}[\sum_{l = L_m}^{L_k} max(0, 1 - \frac{||q_{il} - q_{il-}||_2}{margin + ||q_{il} - q_{il+}||_2})]
%	\end{split}
	\end{eqnarray}}
	\textcolor{ColorName}{where $L_m$ and $L_k$ refer to the $m^{th}$ and $k^{th}$ layers of $E$, respectively. Note that $ margin $ of layer $ l $ is a constant but might vary for different layers.}
	
	\textcolor{ColorName}{Unlike using the concatenated features of RGB images and depth maps} for triplet loss in \cite{piasco2019learning}, we only use features from the shared encoder $E$ to construct the triplet loss, where geometric and semantic information has been fused in $E$ implicitly. This shows that the multi-scale features from $E$ have already contained the information of input image, depth map and segmentation map, indicating that the model has enough ability of recognizing different places through these features. Therefore, there is no need to construct the triplet loss by extracting features from generated depth map and segmentation map explicitly. The effectiveness of multi-scale fusion triplet loss is validated  in Section. \ref{ablation_section}. 
	
		\begin{figure}[thpb]
		\centering
		%\framebox
		%{
		\includegraphics[scale=0.3]{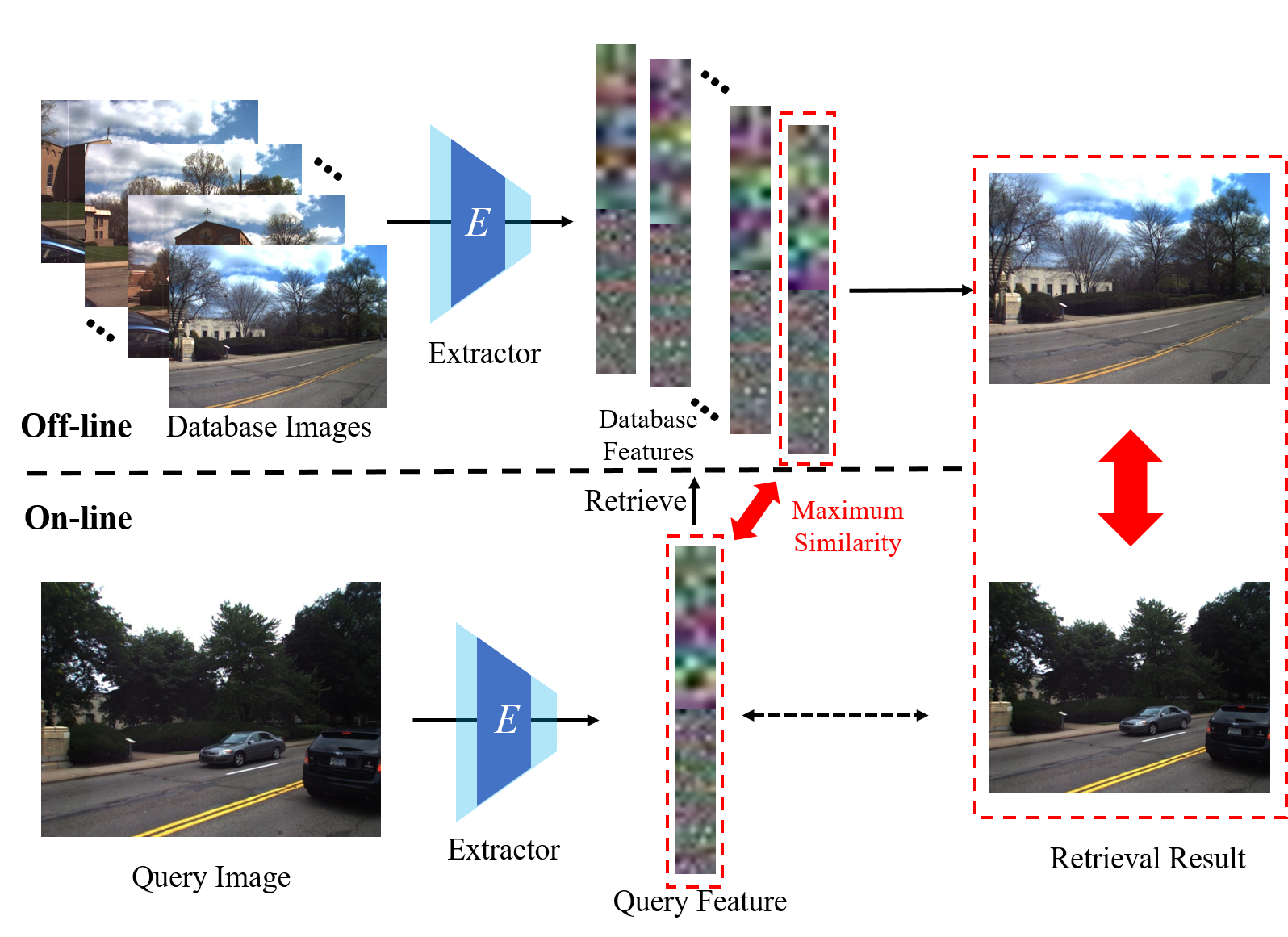}
		%	\vspace{-0.5cm}
		%}
		
		\caption
		{
			The retrieval pipeline is illustrated above. The database images are pre-extracted to build a multi-scale feature database. For the   multi-scale feature extracted from the query image, the database feature with the most similarity or least distance would be retrieved as the result of the corresponded database images. 
		}
		%	\vspace{-0.5cm}
		\label{retrievalpipe}
	\end{figure}
	
	\subsubsection{\textcolor{ColorName}{\textbf{Total Adversarial Training Loss}}}
	 In order to simultaneously train the domain adaptation for semantic segmentation,  depth prediction, and metric learning, all the losses are combined to be a total loss shown in \ref{totalloss}, weighted by $\lambda_T,  \lambda_D,\lambda_S $ respectively.
	 \textcolor{ColorName}{
	 \begin{eqnarray}
	 \label{dis_opt}
	 &\mathop {\min }\limits_D {{\cal L}_{Dis}}\\
	 \label{totalloss}
	 &\mathop {\min }\limits_G {{\cal L}_{Gen} + \lambda_T\mathcal{L}_{mul-T} + \lambda_D\mathcal{L}_{D} + \lambda_S\mathcal{L}_{S} }
	 \end{eqnarray}}
    \textcolor{ColorName}{Note that the GAN loss actually contains two optimizing processes, following the adversarial training pipeline. Other losses, Equation. \ref{multi-scale_depth}, Equation. \ref{multi-scale_seg}, Equation. \ref{multi_triploss} are added in the generation optimizing process \ref{totalloss}, while only discrimination loss in the GAN loss \textcolor{ColorName}{Equation. \ref{GAN-D}} is involved in the discrimination optimizing process \ref{dis_opt}.}
	
 	\subsection{Image Retrieval for Localization}
 	\label{mean_cos}

 	For image-based localization procedure, the feature representations of database should be built first.  For each query image, we will find the one in the database with the most similarity as the image retrieval result, as Figure. \ref{retrievalpipe} shows.
 	
 	As for the representation of database, every image in the database goes through Fusion Feature Extractor $E$ to extract multi-scale fusion feature $f_{db}$, resulting in the fusion features $F_{db}$ of all images in the database is built. For the query image, the same procedure is done to obtain the multi-scale fusion feature ($f_q$).
 	
 	To measure the similarity between $f_q$ and $f_{db}$ as well as taking different scales of features into consideration, we apply $L1$ measure metric \textcolor{ColorName}{or $cosin$ similarity} to find the least distance for retrieval.	Note that the measurement metric used in the triplet loss is $L2$, which is stricter to train the model to learning the best representation.
% Mean Cosine Similarity for all scales of the features and add them together to get the final similarity distance, as shown in the formula \ref{$cosin$}.
 	
%  	%% 换成mean $cosin$的公式, 要体现各个scale求和, 以及体现对channel求和再求平均
%  	\begin{equation}
% 	\begin{split}
% 	\mathcal{S}_{q, db} = \sum\limits_{i} \frac{\sum\limits_{j = 1}^{C} f_{qj} \times f_{dbj}}{\sqrt{\sum\limits_{j = 1}^{C} f_{qj}^2} \times \sqrt{\sum\limits_{j = 1}^{C} f_{dbj}^2} }
% 	\end{split}
% 	\label{$cosin$}
% 	\end{equation}
% 	where $C$ refers to the channel of features and $i$ refers to the $i^{th}$ layer of the Fusion Feature Extractor $E$.
	
% 	The $cosin$ similarity is self-normalized and is suitable for image retreival or place recogniation. Instead of flattening $f_{db}$ and $f_q$ to calculate the traditional $cosin$ similarity along all the channels and pixels, the Mean Cosine Similarity focuses more on the pixel-level similarity for each channel by calculating the mean value of the pixel similarity along the channel dimension. We also validate the effectiveness of mean $cosin$ similarity in Section. \ref{ablation_section}.

	Due to the previous work \cite{sunderhauf2015performance} where the mid-layer feature is the most suitable for image representation for place recognition,  instead of using features extracted from all the layers of the extractor $E$, we only apply  the ones from middle layers of the extractor. The middle-layer features combine the advantages of both the deep and shallow features together, maximally instruct the model to measure the similarity between two images. According to \cite{sunderhauf2015performance}, lower-level feature suffers from the change of view point during the retrieval, while higher-level feature is sensitive to the environmental variance, \textit{i.e.} seasonal change, illumination change, \textit{etc.} Therefore, the middle-layer features are chosen to calculate the similarity finally.
	
	As for image-based localization task, for every query image, the database image which has the least $L1$ distance \textcolor{ColorName}{ or largest $cosin$ similarity} summed by multi-scale features to the query is considered to be the retrieval result. After traversing all the images in the query set, we finally obtain the image-based localization result. 
 	 
	\section{Experiments}
	\label{expe}
	This section introduces the experiments, including the dataset introduction, implementation details, experiment results and ablation studies. 
	\textcolor{ColorName}{Our code and pretrained models are available at \url{https://github.com/HanjiangHu/DASGIL}.}
	
	\subsection{Experimental Setup}

	\subsubsection{\textcolor{ColorName}{\textbf{Datasets for Training}}}
	
	KITTI\cite{geiger2013vision} dataset captures rural areas and roadway with multiple outdoor objects in the scenes by driving around the mid-size city of Karlsruhe. \textcolor{ColorName}{The involved KITTI images for adversarial training are selected according to the setting in \cite{zhao2019geometry}, which is of similar scenes for Virtual KITTI2 dataset.} However, the ground truths of depth map and semantic segmentation map is not accurate and hard to obtain from original Lidar Scanner and camera devices. 
	
	Virtual KITTI 2 \cite{gaidon2016virtual, cabon2020virtual} dataset is a synthetic dataset including five different sequences cloned from real-world KITTI dataset with different camera angle-views (\textit{15 degree left, 15 degree right, 30 degree left, 30 degree right}) and weather conditions (\textit{clone, fog, morning, overcast, rain, sunset}). 	Besides, it contains high-quality ground truths of depth map and semantic segmentation map without any human effort. Due to the multi-conditional image sequences and large-quantity ground truths with high quality of Virtual KITTI 2, the representation  learned from the multi-task model could be well trained on it together with real-world images in KITTI dataset. \textcolor{ColorName}{Note that since the too much change of perspective makes the positive pairs dissimilar, we omit the pairs of angle changes over 30 degrees to prevent the noisy input pairs, like \textit{15 degree left} and \textit{30 degree right}.}
	
	\subsubsection{\textcolor{ColorName}{\textbf{Datasets for Testing}}}
	
	The Extended CMU Seasons dataset \cite{sattler2018benchmarking} is derived from the CMU Visual localization dataset\cite{Badino2011}. It is recorded  over a year along a 9 kilometers long route including urban, suburban, and park areas in Pittsburgh, USA. The left and right images are captured from cameras on both sides of a car. The dataset has 11 query environments and 1 reference environment and is challenging for the huge change of foliage. \textcolor{ColorName}{ The database images are \textit{sunny + No Foliage} while other query images are under the combination of \textit{Cloudy, Overcast, Low Sun, Sunny, Snow} and \textit{Foliage, Mixed Foliage, No Foliage} under all urban, suburban and park areas.}
	
	\textcolor{ColorName}{We have conducted the second series of experiments on Oxford RobotCar dataset \cite{RobotCarDatasetIJRR}. The image dataset was collected from a Bumblebee  camera and three Point Grey Grasshopper2 cameras on the left, rear and right side across one year in Oxford, U.K., including all the weather, season and illumination conditions. Following the setting of recent state-of-the-art work by Piasco \textit{et al.} \cite{piasco2020improving}, we use 1688 images for the reference images with an interval of 5m along 2km. The three query sets are under the similar condition after 7 months (Long-term), snowy weather (Snow) and night scenarios (Night) with about 1000 images respectively.}

	\subsubsection{\textbf{Evaluation Metrics}}
	For the experiments on the Extended  CMU Seasons dataset, we follow the evaluation method on the benchmark website \cite{sattler2018benchmarking} and use its measurement metrics to evaluate the performance of image-retrieval localization. The benchmark on the evaluation website is for visual localization with three levels of precision:  high, medium and coarse precision, \textit{i.e.} ($0.25m,2^\circ$), ($0.5m,5^\circ$) and ($5m,10^\circ$), respectively. The percentage of pose error within each precision is counted to evaluate the performance. The metric we adopt to retrieve the target database images is $L1$.
	
	\textcolor{ColorName}{For the consideration of place recognition in large-scale scenarios, we use recall related metrics for performance evaluation. Similar to  \cite{piasco2020improving}, \textit{Recall @N} is the average ratio of one of the top N retrieved candidates lies within 25m around the groundtruth. \textit{Top-1 Recall @D} is the percentage that the top 1 retrieval lies within distance threshold D form 15m to 50m with respect to the groundtruth position. The retrieval metric in this experiment is $cosin$ distance.}

	\subsubsection{\textbf{Baselines of Image-based Localization}}
	In the experiment, we choose the several state-of-the-art image retrieval-based localization baselines as follows. 
	\begin{itemize}
	    \item \textbf{NetVLAD} \cite{arandjelovic2016netvlad } extracts deep features in VLAD-like networks and uses these to retrieve target images.
	    \item \textbf{DenseVLAD} \cite{torii201524} uses multi-scale dense VLAD of SIFT descriptors for image retrieval in a traditional manner.
	    \item \textbf{DIFL} \cite{hu2019retrieval} learns the domain-invariant feature as representation for retrieval through a self-supervised image-to-image translation architecture.
	    \item \textbf{Xin \textit{et al.}} \cite{xin2019localizing} proposes a Landmark Localization Network to localize the discriminative visual areas that benefit the similarity measurement, which gives the best results currently.
	    \item \textbf{WASABI} \cite{benbihi2020image} retrieves images from the semantic edge wavelet transform through a global image description with the semantic and topological information.
	    \textcolor{ColorName}{\item \textbf{Piasco \textit{et al.}} \cite{piasco2020improving} provides the concatenated feature from RGB and the predicted depth map RGB(D) through Resnet together with the baselines of Resnet RGB-only and Alexnet RGB(H), which are without depth information and hallucination network with depth map respectively.}
	\end{itemize}

	\subsection{Implementation Details}
	
	The virtual images for depth prediction, semantic segmentation and triplet metric learning are from the virtual KITTI 2 dataset, while both of virtual and real-world KITTI images are involved in the GAN loss for domain adaptation. For the retrieval-based localization, the test dataset is the Extended CMU Seasons dataset. Therefore, different datasets are chosen for training and testing respectively and the generalization ability is validated as well across multiple datasets.
	
	% 整体训练的细节写在这
	The original RGB images as well as the ground truths of depth map and segmentation maps are cropped to $256 \times 1024$.  \textcolor{ColorName}{The batch size is set as 8 and learning rate is 0.005 for ADAM optimization algorithm. For the model with the flatten discriminator, the total epoch is set to be 5 to avoid the collapse of adversarial training. While the total training epoch is 40 for the model with the cascade discriminator.}
	
	% depth 和segmentation 网络以及损失，包括用了哪些层，skip connection和构建loss分别用了哪几层
	\textcolor{ColorName}{The RGB images are input into an eight-layer encoder and two decoders for the generation of depth map and segmentation map. Skip connection is applied on all eight layers while the  Multi-scale Depth Reconstruction Loss only involve the last four layers. The Cross Entropy Segmentation Loss only uses the output from the last layer.}

		% GAN 的细节写在这
	\textcolor{ColorName}{For the domain adaptation at the feature level, the Flatten Discriminator (FD) is a three-layer fully linear neural network with  dimension as 1004800, 64, 64, 1. 	Due to the eight-layer skip connection structure, the multi-layer features from the encoder are flattened and concatenated, going through the a BatchNorm layer right before fed to discriminator. For the Cascade Discriminator (CD), the input channel of each ResBlock is exactly consistent with the channel of last feature map while the output channel is the same as that of next feature map. The linear layer with dimension of 1536 is attached at the end to regress the value for true or false label.}
	\begin{table}[]
		\caption{Results in Different Region Environments}
		\centering
		\label{overview_results}
		\begin{tabular}{c|c|c|c}
			\hline
			\textbf{Methods} & \begin{tabular}[c]{@{}c@{}}\textbf{Park}\\ .25m/0.5m/5m\\ 2$^{\circ}$/5$^{\circ}$/10$^{\circ}$\end{tabular} & \begin{tabular}[c]{@{}c@{}}\textbf{Suburban}\\ .25m/0.5m/5m\\ 2$^{\circ}$/5$^{\circ}$/10$^{\circ}$\end{tabular} & \begin{tabular}[c]{@{}c@{}}\textbf{Urban}\\ .25m/0.5m/5m\\ 2$^{\circ}$/5$^{\circ}$/10$^{\circ}$\end{tabular} \\ \hline
			NetVLAD\cite{arandjelovic2016netvlad} & 2.6/  10.4/ 55.9 & 3.7/ 13.9/ 74.7 & 12.2/ 31.5/ 89.8 \\
			DenseVLAD\cite{torii201524} & 5.2/ 19.1/ 62.0 & 5.3/ 18.7/ 73.9 & 14.7/ 36.3/ 83.9 \\
			DIFL\cite{hu2019retrieval} & 6.1/ 20.7/ 69.1 & 5.6/ 18.2/ 69.8 & 14.8/ 35.1/ 79.6   \\
			Xin \textit{et al.}\cite{xin2019localizing} & 6.6/ 23.1/ 73.0 & 5.8/ 19.4/ 76.1 & 17.3/ \textbf{42.5}/ 89.0 \\ 
			WASABI\cite{benbihi2020image} & 2.4/ 9.1 / 54.5 & 3.8/ 13.9/ 67.3 & 7.9 / 21.3/ 75.2   \\ \hline
			\textcolor{ColorName}{\textbf{DASGIL(CD)}} & \textcolor{ColorName}{7.5/ 25.3/ 79.2} & \textcolor{ColorName}{6.5/ 21.5/ 86.5} & \textcolor{ColorName}{17.3/ 42.2/ 90.5} \\
			\textcolor{ColorName}{\textbf{DASGIL(FD)}} & \textcolor{ColorName}{\textbf{7.9}/ \textbf{26.9}/ \textbf{83.5}} & \textcolor{ColorName}{\textbf{6.7}/ \textbf{22.1}/ \textbf{88.5}} & \textcolor{ColorName}{\textbf{17.4}/ 42.0/ \textbf{91.1}} \\ \hline
		\end{tabular}
	\end{table}
	
	\begin{table}[]
		\caption{Results of Different \textbf{Vegetation} Conditions \protect\\ Database	Reference is with \textbf{No Foliage}}
		\centering
		\label{foliage_results}
		\begin{tabular}{c|c|c}
			\hline
			\textbf{Methods} & \begin{tabular}[c]{@{}c@{}}\textbf{Foliage}\\ 0.25m / 0.5m / 5m\\ 2$^{\circ}$ / 5$^{\circ}$ / 10$^{\circ}$\end{tabular} & \begin{tabular}[c]{@{}c@{}}\textbf{Mixed Foliage}\\ 0.25m / 0.5m / 5m\\ 2$^{\circ}$ / 5$^{\circ}$ / 10$^{\circ}$\end{tabular}  \\ \hline
			NetVLAD\cite{arandjelovic2016netvlad} & 6.2 / 18.5 / 74.3 & 5.8 / 17.6 / 71.1  \\
			DenseVLAD\cite{torii201524} & 7.4 / 21.1 / 68.0 & 8.5 / 24.5 / 73.0  \\
			DIFL\cite{hu2019retrieval} & 8.2 / 22.2 / 69.0 & 9.6 / 26.0 / 74.4\\
			Xin \textit{et al.}\cite{xin2019localizing} & 9.5 / 26.7 / 77.4 & 10.3 / 28.4 / 79.0 \\ 
			WASABI\cite{benbihi2020image} & 4.9 / 15.2 / 67.6 & 4.8 / 14.8 / 64.9  \\ \hline
			\textcolor{ColorName}{\textbf{DASGIL(CD)}} & \textcolor{ColorName}{10.1 / 28.0 / 82.5} & \textcolor{ColorName}{11.3 / 31.0 / 87.8}  \\ 
			\textcolor{ColorName}{\textbf{DASGIL(FD)}} & \textcolor{ColorName}{\textbf{10.4} / \textbf{28.7} / \textbf{84.3}} & \textcolor{ColorName}{\textbf{11.4} / \textbf{31.5} / \textbf{91.2}}  \\ \hline
		\end{tabular}
	\end{table}
	
	% triplet 的细节写在这，包括用了哪些层，训练哪几层，测试哪几层
	\begin{table*}[]
		\centering
		\caption{Results of Different \textbf{Weather} Conditions \protect\\Database Reference is under \textbf{Sunny}}
		\centering
		\label{weather_results}
		\begin{tabular}{c|c|c|c|c}
			\hline
			\textbf{Methods} & \begin{tabular}[c]{@{}c@{}}\textbf{Overcast}\\ 0.25m / 0.5m / 5m\\ 2$^{\circ}$ / 5$^{\circ}$ / 10$^{\circ}$\end{tabular} & \begin{tabular}[c]{@{}c@{}}\textbf{Low Sun}\\ 0.25m / 0.5m / 5m\\ 2$^{\circ}$ / 5$^{\circ}$ / 10$^{\circ}$\end{tabular} & \begin{tabular}[c]{@{}c@{}}\textbf{Cloudy}\\ 0.25m / 0.5m / 5m\\ 2$^{\circ}$ / 5$^{\circ}$ / 10$^{\circ}$\end{tabular} & \begin{tabular}[c]{@{}c@{}}\textbf{Snow}\\ 0.25m / 0.5m / 5m\\ 2$^{\circ}$ / 5$^{\circ}$ / 10$^{\circ}$\end{tabular}  \\ \hline
			NetVLAD\cite{arandjelovic2016netvlad} & 6.7 / 19.1 / 76.3  & 5.5 / 17.5 / 71.3 & 6.8 / 20.1 / 79.5 & 5.0 / 16.4 / 68.0  \\
			DenseVLAD\cite{torii201524}  & 8.4 / 23.3 / 72.1  & 8.3 / 26.1 / 76.0  & 9.3 / 27.3 / 80.5 & \textbf{8.3} / 29.0 / 78.9 \\
			DIFL\cite{hu2019retrieval} &9.7 / 25.3 / 70.9  & 8.7 / 25.3 / 74.4 & 8.8 / 24.7 / 76.9 & 7.4 / 26.7 / 73.5 \\
			Xin \textit{et al.}\cite{xin2019localizing} & 11.5 / 30.8 / 80.8  & 9.3 / 27.6 / 79.2  & 9.4 / 28.0 / 83.7 & 7.6 / 27.6 / 75.9\\ 
			WASABI\cite{benbihi2020image} & 5.4 / 15.8 / 70.8 & 4.2 / 14.0 / 62.1 & 5.1 / 15.3 / 71.0   & 3.4 / 13.2 / 58.0 \\ \hline
			\textcolor{ColorName}{\textbf{DASGIL(CD)}} & \textcolor{ColorName}{\textbf{12.7} / \textbf{32.5} / 86.3} &  \textcolor{ColorName}{9.8 / 29.5 / 87.3} & \textcolor{ColorName}{9.8 / 28.6 / 89.8}  & \textcolor{ColorName}{7.7 / 29.3 / 85.3} \\ 
			\textcolor{ColorName}{\textbf{DASGIL(FD)}} & \textcolor{ColorName}{12.5 / 32.4 / \textbf{88.6}} &  \textcolor{ColorName}{\textbf{10.1} / \textbf{30.3} / \textbf{89.6}} & \textcolor{ColorName}{\textbf{10.0} / \textbf{28.9} / \textbf{92.1}}  & \textcolor{ColorName}{8.1 / \textbf{30.1} / \textbf{86.6}} \\ \hline	
		\end{tabular}
	\end{table*}
\begin{figure*}[thpb]
	\centering
	
	\begin{mdframed}[hidealllines=true]%,backgroundcolor=blue!20]%,
		%\lipsum[2]
		\includegraphics[scale=0.35]{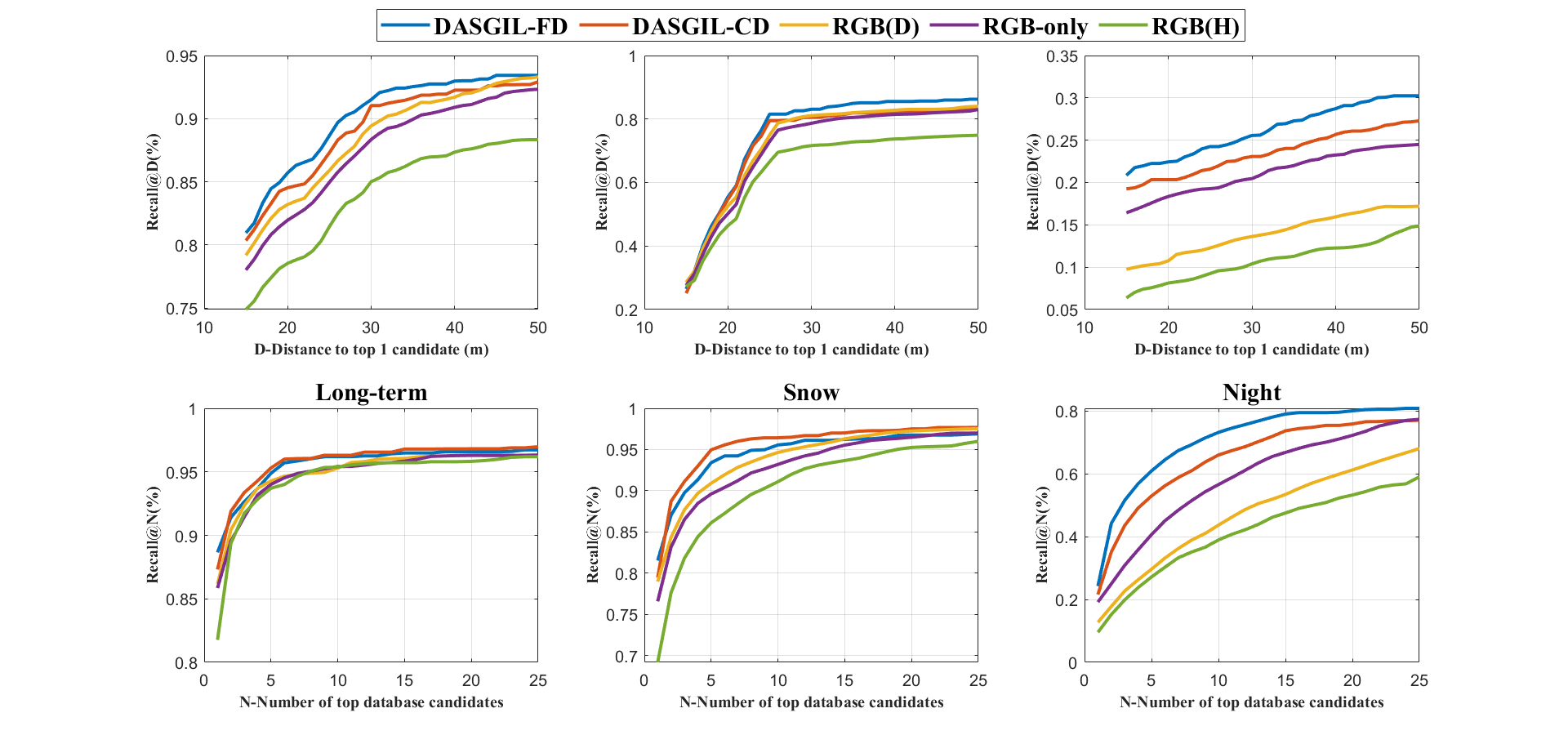}
	\end{mdframed}	
	\caption
	{
		\textcolor{ColorName}{The comparison results on Oxford RobotCar dataset are shown above with baselines of RGB(D), RGB-only and RGB(H). The first row is the top 1 recall of different distances while the second row is the top N recall from N retrieved candidates. The columns form left to right are with the query of Long-term, Snow and Night respectively. The results of baselines are from the previous work \cite{piasco2020improving}.}
	}
	%	\vspace{-0.5cm}
	\label{robotcar_result}
\end{figure*}
	
	Since there are different variations on camera angles and environments for any image sequence in the virtual KITTI 2 dataset, the positively paired images are within 5-image-interval along a sequence but from different environments while the negatively paired ones are randomly chosen and flipped. \textcolor{ColorName}{The negative pairs could come from all the environments with different scenes. All the images are randomly horizontally flipped for data augmentation while keeping the same transformation for positive pairs. The features from the middle-four layers are used to construct the multi-scale triplet loss. For image retrieval, features from the fifth and sixth layers are used as representation with FD while only the fifth layer feature is used for CD model.} The margins are set to be 1 for the features at the third, fourth, fifth and sixth layer in the multi-scale triplet loss.

	\subsection{\textcolor{ColorName}{Extended CMU-Seasons Dataset Results}}
	We conduct a series of experiments under different regional environments, vegetation conditions and weather conditions. The results of other baselines are from the long-term visualization benchmark website \url{https://www.visuallocalization.net/benchmark}.
	\subsubsection{\textbf{Results in Different Regional  Environments}}
	\label{environment}
	
	We have validated the effectiveness of our model under various regional  environments, as shown in Table. \ref{overview_results}. Due to the variety of regional   environments in the real world, \textit{i.e.} urban area, suburban area, and other areas, it is of great necessity to validate the effectiveness of the model in various areas.
%	 Besides, the scenario of different environments varies significantly. For example, in the urban area, there exist lots of cars, roads and modern buildings while there are not many trees. In park area, however, trees occupy most of the image but there are almost no buildings and cars. Therefore, the model should be robust for place recognition in these different areas. 

\textcolor{ColorName}{From Table. \ref{overview_results}, we can conclude that our model with Flatten Discriminator behaves almost the best among all of the baselines under three regional conditions: urban, suburban and park. The model with Cascade Discriminator also outperforms other baselines in most cases. Specifically, in suburban area, our model performs 12.4\% higher than state-of-the-art  baselines under the coarse precision, which is much more important than that in high precision because there is numerous methods for finer pose regression for localization.The performance in park and suburban indicates that the geometric and semantic is more applicable to trees and other static objects.}

As for urban area, our model performs the best except under medium precision. Because in the urban area there are too many cars and other dynamic objects  which could move across environments, geometric and semantic information would be affected and not consistent even for the same place under changing environments. These factors make our model perform not as satisfactorily as in suburban area or park area.
% However, for coarse-precision localization, our model is robust to dynamic objects, making ours the best compared with other baselines. 

\begin{table*}[]
	\caption{Ablation Study: \textcolor{ColorName}{Training} Modules in DASGIL}
	\label{ablation1}
	\centering
	\begin{tabular}{|ccccccc|c|c|c|}
		\hline
		Depth        & Segmentation & \begin{tabular}[c]{@{}c@{}}\textcolor{ColorName}{Single}\\ \textcolor{ColorName}{FD}\end{tabular} & \begin{tabular}[c]{@{}c@{}}\textcolor{ColorName}{Multiple}\\ \textcolor{ColorName}{FD}\end{tabular} & \begin{tabular}[c]{@{}c@{}}\textcolor{ColorName}{Multiple}\\ \textcolor{ColorName}{CD}\end{tabular} & \begin{tabular}[c]{@{}c@{}}Single\\ Triplet\\ Loss\end{tabular} & \begin{tabular}[c]{@{}c@{}}Multiple\\ Triplet\\ Loss\end{tabular} & \begin{tabular}[c]{@{}c@{}}\textbf{Park}\\ 0.25m / 0.5m / 5m\\ 2$^{\circ}$ / 5$^{\circ}$ / 10$^{\circ}$\end{tabular} & \begin{tabular}[c]{@{}c@{}}\textbf{Suburban}\\ 0.25m / 0.5m / 5m\\ 2$^{\circ}$ / 5$^{\circ}$ / 10$^{\circ}$\end{tabular} & \begin{tabular}[c]{@{}c@{}}\textbf{Urban}\\ 0.25m / 0.5m / 5m\\ 2$^{\circ}$ / 5$^{\circ}$ / 10$^{\circ}$\end{tabular} \\ \hline
		$\checkmark$ & $\times$     & $\times$                                            & $\checkmark$                                          & \textcolor{ColorName}{$\times$}                                              & $\times$                                                        & $\checkmark$                                                      & \textcolor{ColorName}{7.1} / \textcolor{ColorName}{24.7} / \textcolor{ColorName}{78.9}                                                                                           & \textcolor{ColorName}{6.6} / \textcolor{ColorName}{21.9} / \textcolor{ColorName}{86.0}                                                                                               & \textcolor{ColorName}{16.9} / \textcolor{ColorName}{40.7} / \textcolor{ColorName}{88.8}                                                                                           \\
		$\times$     & $\checkmark$ & $\times$                                            & $\checkmark$                                          & \textcolor{ColorName}{$\times$}                                              & $\times$                                                        & $\checkmark$                                                      & \textcolor{ColorName}{6.9} / \textcolor{ColorName}{23.6} / \textcolor{ColorName}{74.1}                                                                                           & \textcolor{ColorName}{6.5} / \textcolor{ColorName}{21.4} / \textcolor{ColorName}{81.9}                                                                                               & \textcolor{ColorName}{16.0} / \textcolor{ColorName}{38.8} / \textcolor{ColorName}{86.1}                                                                                           \\
		$\checkmark$ & $\checkmark$ & $\times$                                            & $\times$                                              & \textcolor{ColorName}{$\times$}                                              & $\times$                                                        & $\checkmark$                                                      & \textcolor{ColorName}{4.8} / \textcolor{ColorName}{16.2} / \textcolor{ColorName}{57.8}                                                                                           & \textcolor{ColorName}{4.9} / \textcolor{ColorName}{16.2} / \textcolor{ColorName}{64.7}                                                                                               & \textcolor{ColorName}{12.9} / \textcolor{ColorName}{31.7} / \textcolor{ColorName}{76.5}                                                                                           \\
		$\checkmark$ & $\checkmark$ & $\checkmark$                                        & $\times$                                              & \textcolor{ColorName}{$\times$}                                              & $\times$                                                        & $\checkmark$                                                      & \textcolor{ColorName}{6.3} / \textcolor{ColorName}{23.6} / \textcolor{ColorName}{75.3}                                                                                          & \textcolor{ColorName}{5.9} / \textcolor{ColorName}{20.1} / \textcolor{ColorName}{81.7}                                                                                               & \textcolor{ColorName}{16.1} / \textcolor{ColorName}{40.0} / \textcolor{ColorName}{85.8}                                                                                           \\
		\textcolor{ColorName}{$\checkmark$} & \textcolor{ColorName}{$\checkmark$} & \textcolor{ColorName}{$\times$}                                            & \textcolor{ColorName}{$\times$}                                              & \textcolor{ColorName}{$\checkmark$}                                          & \textcolor{ColorName}{$\times$}                                                        & \textcolor{ColorName}{$\checkmark$}                                                      & \textcolor{ColorName}{7.5} / \textcolor{ColorName}{25.3} / \textcolor{ColorName}{79.2}                                                                                           & \textcolor{ColorName}{6.5} / \textcolor{ColorName}{21.5} / \textcolor{ColorName}{86.5}                                                                                               & \textcolor{ColorName}{17.3} / \textcolor{ColorName}{42.2} / \textcolor{ColorName}{90.5}                                                                                           \\
		$\checkmark$ & $\checkmark$ & $\times$                                            & $\checkmark$                                          & \textcolor{ColorName}{$\times$}                                              & $\checkmark$                                                    & $\times$                                                          & \textcolor{ColorName}{7.2} / \textcolor{ColorName}{24.8} / \textcolor{ColorName}{78.8}                                                                                           & \textcolor{ColorName}{6.4} / \textcolor{ColorName}{21.4} / \textcolor{ColorName}{85.7}                                                                                               & \textcolor{ColorName}{17.0} / \textcolor{ColorName}{40.9} / \textcolor{ColorName}{89.3}                                                                                           \\
		$\checkmark$ & $\checkmark$ & $\times$                                            & $\checkmark$                                          & \textcolor{ColorName}{$\times$}                                              & $\times$                                                        & $\checkmark$                                                      & \textcolor{ColorName}{\textbf{7.9}} / \textcolor{ColorName}{\textbf{26.9}} / \textcolor{ColorName}{\textbf{83.5}}                                                                                           & \textcolor{ColorName}{\textbf{6.7}} / \textcolor{ColorName}{\textbf{22.1}} / \textcolor{ColorName}{\textbf{88.5}}                                                                                               & \textcolor{ColorName}{\textbf{17.4}} / \textcolor{ColorName}{\textbf{42.0}} / \textcolor{ColorName}{\textbf{91.1}}                                                                                        \\ \hline
	\end{tabular}
\end{table*}

\begin{table*}[]
	\caption{\textcolor{ColorName}{Ablation Study: Feature Representation for Retrieval}}
	\label{ablation2}
	\centering
	\begin{tabular}{|c|cccc|c|c|c|}
		\hline
		\textcolor{ColorName}{Discriminator}& \textcolor{ColorName}{Single-layer} & \textcolor{ColorName}{Multi-layer}  & \textcolor{ColorName}{NetVLAD}      & \textcolor{ColorName}{R-MAC}        & \begin{tabular}[c]{@{}c@{}}\textcolor{ColorName}{\textbf{Park}}\\ \textcolor{ColorName}{0.25m / 0.5m / 5m}\\ \textcolor{ColorName}{2$^{\circ}$ / 5$^{\circ}$ / 10$^{\circ}$}\end{tabular} & \begin{tabular}[c]{@{}c@{}}\textcolor{ColorName}{\textbf{Suburban}}\\ \textcolor{ColorName}{0.25m / 0.5m / 5m}\\ \textcolor{ColorName}{2$^{\circ}$ / 5$^{\circ}$ / 10$^{\circ}$}\end{tabular} & \begin{tabular}[c]{@{}c@{}}\textcolor{ColorName}{\textbf{Urban}}\\ \textcolor{ColorName}{0.25m / 0.5m / 5m}\\ \textcolor{ColorName}{2$^{\circ}$ / 5$^{\circ}$ / 10$^{\circ}$}\end{tabular} \\ \hline
		\multirow{4}{*}{\textcolor{ColorName}{Flatten}} & \textcolor{ColorName}{$\checkmark$} & \textcolor{ColorName}{$\times$}     & \textcolor{ColorName}{$\times$}     & \textcolor{ColorName}{$\times$}     & \textcolor{ColorName}{7.6 / 26.2 / 83.2}                                                                                           & \textcolor{ColorName}{6.5 / 21.2 / 87.5}                                                                                               & \textcolor{ColorName}{17.2 / 41.4 / 90.9}                                                                                           \\
		& \textcolor{ColorName}{$\times$}     & \textcolor{ColorName}{$\checkmark$} & \textcolor{ColorName}{$\checkmark$} & \textcolor{ColorName}{$\times$}     & \textcolor{ColorName}{0.9 / 3.5 / 25.8}                                                                                            & \textcolor{ColorName}{1.7 / 5.7 / 39.3}                                                                                                & \textcolor{ColorName}{5.5 / 14.4 / 54.1}                                                                                            \\
		& \textcolor{ColorName}{$\times$}     & \textcolor{ColorName}{$\checkmark$} & \textcolor{ColorName}{$\times$}     & \textcolor{ColorName}{$\checkmark$} & \textcolor{ColorName}{1.4 / 4.9 / 29.9}                                                                                            & \textcolor{ColorName}{1.9 / 6.7 / 37.9}                                                                                                & \textcolor{ColorName}{5.5 / 13.7 / 52.1}                                                                                            \\
		& \textcolor{ColorName}{$\times$}     & \textcolor{ColorName}{$\checkmark$} & \textcolor{ColorName}{$\times$}     & \textcolor{ColorName}{$\times$}     & \textcolor{ColorName}{\textbf{7.9 / 26.9 / 83.5}}                                                                                           & \textcolor{ColorName}{\textbf{6.7 / 22.1 / 88.5}}                                                                                               & \textcolor{ColorName}{\textbf{17.4 / 42.0 / 91.1}}                                                                                           \\ \hline
		\multirow{4}{*}{\textcolor{ColorName}{Cascade}} & \textcolor{ColorName}{$\checkmark$} & \textcolor{ColorName}{$\times$}     & \textcolor{ColorName}{$\times$}     & \textcolor{ColorName}{$\times$}     & \textcolor{ColorName}{\textbf{7.5 / 25.3 / 79.2}}                                                                                           & \textcolor{ColorName}{\textbf{6.5} / 21.5 / 86.5}                                                                                               & \textcolor{ColorName}{\textbf{17.3 / 42.2 / 90.5}}                                                                                           \\
		& \textcolor{ColorName}{$\times$}     & \textcolor{ColorName}{$\checkmark$} & \textcolor{ColorName}{$\checkmark$} & \textcolor{ColorName}{$\times$}     & \textcolor{ColorName}{0.9 / 3.0 / 27.0}                                                                                            & \textcolor{ColorName}{1.0 / 3.9 / 33.5}                                                                                                & \textcolor{ColorName}{2.7 / 7.8 / 43.6}                                                                                             \\
		& \textcolor{ColorName}{$\times$}     & \textcolor{ColorName}{$\checkmark$} & \textcolor{ColorName}{$\times$}     & \textcolor{ColorName}{$\checkmark$} & \textcolor{ColorName}{0.7 / 2.5 / 22.8}                                                                                            & \textcolor{ColorName}{1.0 / 3.7 / 27.0}                                                                                                & \textcolor{ColorName}{2.5 / 7.0 / 36.8}                                                                                             \\
		& \textcolor{ColorName}{$\times$}     & \textcolor{ColorName}{$\checkmark$} & \textcolor{ColorName}{$\times$}     & \textcolor{ColorName}{$\times$}     & \textcolor{ColorName}{7.1 / 24.9 / 79.0}                                                                                           & \textcolor{ColorName}{\textbf{6.5 / 21.9 / 86.8}}                                                                                               & \textcolor{ColorName}{17.1 / 41.7 / \textbf{90.5}}                                                                                           \\ \hline
	\end{tabular}

%	\begin{tablenotes}

%	\end{tablenotes}
\end{table*}

\subsubsection{\textbf{Results under Different Vegetation Conditions}}
	
The proposed model is tested under different foliage conditions under all the areas of urban, suburban and park,  and the results are shown in Table. \ref{foliage_results}, with the same evaluation metrics as in Section.\ref{environment}. Since the reference vegetation is \textit{No Foliage}, the results under different foliage conditions indicate the effectiveness of our model under various foliage conditions, including foliage and mixed foliage. \textcolor{ColorName}{Our model with FD performs the best compared to all baselines under these vegetation conditions, resulting in the robustness to the huge change of vegetation.} Among the foliage conditions, our model performs the best when there is mixed foliage in the environment. Due to the geometric and semantic information we  use for this task, our model has robustness to the change of foliage with multi-scale deep features.

	\subsubsection{\textbf{Results under Different Weather Conditions}}
	
	Besides the experiments on the change of regional  environment and vegetation conditions, we also validate our model under different  weather conditions for the images in all the areas, including low sun, cloudy, overcast and snow. Note that the reference images in the database are under sunny condition. The result are shown in Table. \ref{weather_results}, with the same evaluation method as in Section. \ref{environment}. 	\textcolor{ColorName}{Our models with CD and FD outperform the other state-of-the-art baselines under most the weather conditions especially on the coarse precision.} Note that the training set Virtual KITTI 2 does not contain snow condition, but our model still performs satisfactorily  under the snow condition on medium and coarse precision, showing the strong generalization ability of our model.

\subsection{\textcolor{ColorName}{Oxford RobotCar Dataset Results}}

\textcolor{ColorName}{We conduct the experiments with our model of Cascade Discriminator (CD) and Flatten Discriminator (FD), compared to the results from baselines RGB(D), RGB-only and RGB(H) in \cite{piasco2020improving}. The results are shown in Figure. \ref{robotcar_result}}

\subsubsection{\textcolor{ColorName}{\textbf{Results for Top-1 Recall @D}}}
\textcolor{ColorName}{From the first row of Figure. \ref{robotcar_result}, it could be seen that our method with FD performs the best compared to the baselines, especially for Long-term and Night scenarios. For the snow query images, the top 1 recall at a near distance is the same using different algorithms, which indicates that the snowy environment is so challenging that the high-precision place recognition sucks through all the methods. Although for Night scenario the performance improvement of ours is obvious, the top 1 recall is quite low even with the distance threshold of 50m, showing that there is still a long way to go for the day-night image retrieval. Our model with CD perform the second best in most cases under the metric of \textit{Top-1 Recall @D}, which is consistent with the results of CMU-Seasons dataset which are also the top 1 ratio within the specific error thresholds. The results show that our method does well not only in the high-precision localization but also in the large-scale place recognition.}

\subsubsection{\textcolor{ColorName}{\textbf{Results for Recall @N}}}
\textcolor{ColorName}{Figure. \ref{robotcar_result} illustrates the recall@N results on the second row for each query conditions. The results of our models with FD and CD are better than the baselines from \cite{piasco2020improving}. Different from the cases of the top 1 recall, DASGIL-CD performs better than DASGIL-FD for Long-term and Snow place recognition, which shows that the cascade discriminator is beneficial to the recall with more candidates under challenging environments. Although the precision under Night scenarios is low from top-1 recall, the recall is petty satisfactory with more than 10 retrieved candidates, which is of greater significance for the practical outdoor place recognition for the autonomous driving or mobile robots. }

\subsection{Ablation Study}
\label{ablation_section}
The effectiveness of the modules in the architecture of DASGIL as well as the methods of training and testing are validated through a series of comparison experiments in ablation study section. % todo cascade and flatten, multitrip over single trip due to remove 30 degrees improvements
\subsubsection{\textcolor{ColorName}{\textbf{Training Modules in DASGIL}}}

 \textcolor{ColorName}{Table. \ref{ablation1}  shows the impact of different modules in DASGIL, including the generation module of both depth map and segmentation map as well as the types of discriminator of GAN module. Besides, the multi-layer triplet loss is also compared to the loss of single feature map for metric learning. It could be concluded that all of the \textit{Depth} generation module and \textit{Segmentation} generation module are effective and indispensable in the proposed DASGIL framework. The geometric information from the depth map generation is more important to the place recognition than the semantic information form segmentation map.}
 
 \textcolor{ColorName}{While for different types of discriminator, \textit{Single FD}  represents that only single-layer feature is used when training flatten discriminator, where the 5th layer feature is the best among all the 8 layers in the experiment and the retrieval is based on the 5th layer as well. \textit{Multiple FD} and \textit{Multiple CD} use features from all the layers as input. It could be seen that without any discriminator the performance is the worst and the results of multi-layer discriminator are better than single flatten discriminator.
 Also,  \textit{Multiple Triplet Loss} ($3^{rd}$ to $6^{th}$ layers) gives better the results compared to the best \textit{Single Triplet Loss} ($5^{th}$ layer) for triplet loss, confirming the claim in \ref{triplet_loss}.}

\subsubsection{\textcolor{ColorName}{\textbf{Feature Representation for Retrieval}}}

\textcolor{ColorName}{Once the model is well trained using features from multiple layers for metric learning and adversarial training, the multi-scale latent feature could represent the real-world image. However, it might be not efficient to use all the features involved in the training process since they could be affected by noise or dynamic objects for the image retrieval. Besides, other pooling methods like NetVLAD \cite{arandjelovic2016netvlad} and R-MAC \cite{tolias2015particular} are also commonly used for image retrieval. We have conducted another series of experiments to investigate how feature representation is the most suitable for the proposed model and the results as shown in Table. \ref{ablation2}.}

\textcolor{ColorName}{The model used in this ablation study is with multi-layer triplet loss from the $3^{rd}$ to $6^{th}$ layers. The \textit{Single-layer} indicates that the $5^{th}$ layer feature is used for retrieval which is the best among all layers. The \textit{Multi-layer} uses the features from $5^{th}$ and $6^{th}$ layers for best retrieval results. NetVLAD layer is fine-tuned on the fixed pretrained DASGIL models of both FD and CD models for 5 epoch. R-MAC is implemented directly to the features through pretrained models when testing. For the multi-layer feature, we adopt the $L2$ distance to sum up the output of NetVLAD or R-MAC for image retrieval.}

\textcolor{ColorName}{It could be seen from the Table. \ref{ablation2} that NetVLAD and R-MAC perform much worse than the usage of the direct $L1$ distance for the single-layer or multi-layer features, which shows that the mid features for metric learning from pretrained DASGIL are not suitable to represent the deep semantic features as the pretrained classification backbones do, like VGG, ResNet, etc. For the model with Flatten Discriminator, the retrieval with multi-layer feature is better than the single-layer feature. While the model with Cascade Discriminator obtains better results with single-layer feature retrieval in park and urban scenarios, indicating that the Cascade Discriminator module could make the feature from each layer follow the same distribution more strictly so that single-layer representation could be used for image retrieval and localization.}

	\section{Conclusion}
	\label{conc}
	
	In this paper, we propose a novel multi-task architecture, DASGIL,  to fully extract geometric and semantic information for retrieval-based localization. Our method implements domain adaptation from synthetic to real-world images \textcolor{ColorName}{through two novel multi-scale discriminators} and fuses the features from original image, depth maps and semantic segmentation maps. Besides, the experiments are conducted on the Extended CMU Seasons and \textcolor{ColorName}{Oxford RobotCar dataset to validate the performance on image-based localization and large-scale place recognition}, resulting in outperforming state-of-the-art  baselines for retrieval-based localization under changing environments.
\ifCLASSOPTIONcaptionsoff
  \newpage
\fi

	\bibliographystyle{IEEEtran}
    \bibliography{IEEEabrv,mylib}

\begin{IEEEbiography}[{\includegraphics[width=1in,height=1.25in,clip,keepaspectratio]{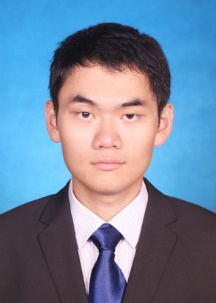}}]{Hanjiang Hu}
	received the B.Eng degree in Mechanical Engineering from Shanghai Jiao Tong University, Shanghai, China, in 2018. He is currently working toward the M.S. degree in Control Science and Engineering at Shanghai Jiao Tong University, Shanghai, China. His current research interests include visual perception and localization, autonomous driving, computer vision and machine learning.
\end{IEEEbiography}
%\vspace{-3cm}

\begin{IEEEbiography}[{\includegraphics[width=1in,height=1.25in,clip,keepaspectratio]{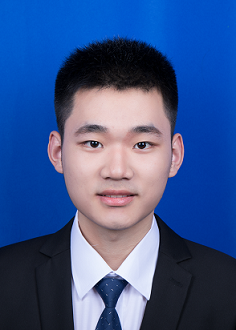}}]{Zhijian Qiao}
received the B.Eng degree in Automation from Northeastern University, Shenyang, China, in 2019. He is currently working toward the M.S. degree in Control Engineering at Shanghai Jiao Tong University, Shanghai, China. His current research interests include visual SLAM, mobile robotics, and point cloud registration.
\end{IEEEbiography}

\begin{IEEEbiography}[{\includegraphics[width=1in,height=1.25in,clip,keepaspectratio]{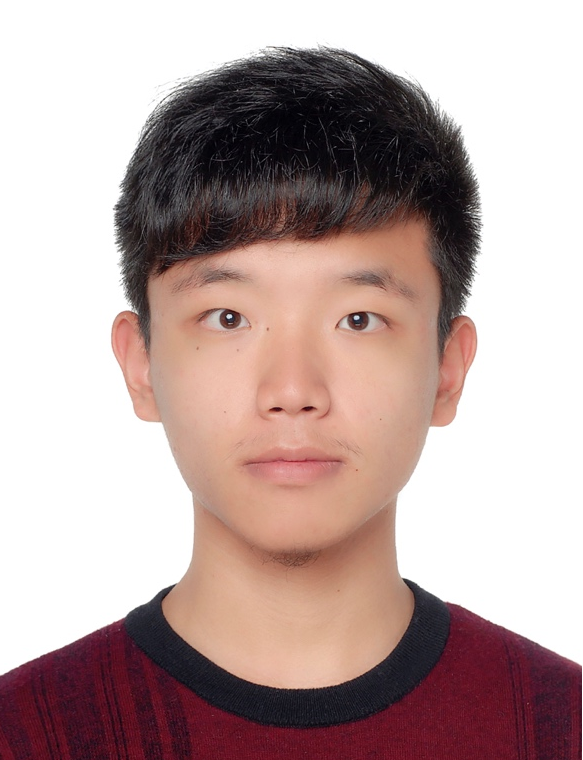}}]{Ming Cheng}
	is currently pursuing the B.S. degree in Department of Automation, Shanghai Jiao Tong University, Shanghai, China. He is now doing research at Autonomous Robot Lab, Shanghai Jiao Tong University. His research interests are deep learning, computer vision and autonomous driving.
\end{IEEEbiography}

\begin{IEEEbiography}[{\includegraphics[width=1in,height=1.25in,clip,keepaspectratio]{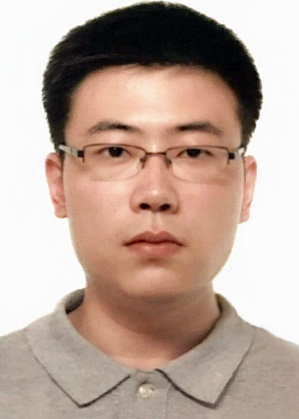}}]{Zhe Liu}
	received his B.S. degree in Automation from Tianjin University, Tianjin, China, in 2010, and Ph.D. degree in Control Technology and Control Engineering from Shanghai Jiao Tong University, Shanghai, China, in 2016. From 2017 to 2020, he was a Post-Doctoral Fellow with the Department of Mechanical and Automation Engineering, The Chinese University of Hong Kong, Hong Kong. He is currently a Research Associate with the Department of Computer Science and Technology, University of Cambridge. His research interests include autonomous mobile robot, multirobot cooperation and autonomous driving system.
\end{IEEEbiography}

\begin{IEEEbiography}[{\includegraphics[width=1in,height=1.25in,clip,keepaspectratio]{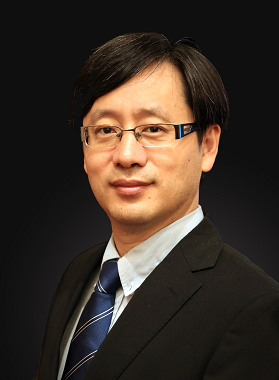}}]{Hesheng Wang}
	(SM’15) received the B.Eng. degree in electrical engineering from the Harbin Institute of Technology, Harbin, China, in 2002, and the M.Phil. and Ph.D. degrees in automation and computer-aided engineering from The Chinese University of Hong Kong, Hong Kong, in 2004 and 2007, respectively. He was a Post-Doctoral Fellow and Research Assistant with the Department of Mechanical and Automation Engineering, The Chinese University of Hong Kong, from 2007 to 2009. He is currently a Professor with the Department of Automation, Shanghai Jiao Tong University, Shanghai, China. His current research interests include visual servoing, service robot, adaptive robot control, and autonomous driving. 
	Dr. Wang is an Associate Editor of Assembly Automation and the International Journal of Humanoid Robotics, a Technical Editor of the IEEE/ASME TRANSACTIONS ON MECHATRONICS. He served as an Associate Editor of the IEEE TRANSACTIONS ON ROBOTICS from 2015 to 2019. He was the General Chair of the IEEE RCAR 2016, and the Program Chair of the IEEE ROBIO 2014 and IEEE/ASME AIM 2019.
\end{IEEEbiography}
%

%\vspace{-15cm}

% \bibliography{IEEEabrv,mylib}
% \bibliographystyle{ieeetr} %ieeetr国际电气电子工程师协会期刊
% \bibliography{mylib} % ref就是之前建立的ref.bib文件的前缀

%\vfill


\end{document}